\def\paperID{9670}
\def\confName{CVPR}
\def\confYear{2025}

\def\paperTitle{A General Adaptive Dual-level Weighting Mechanism for Remote Sensing Pansharpening}

\def\authorBlock{
    Jie Huang\textsuperscript{\rm 1}\thanks{Equal contribution.} \qquad
    Haorui Chen\textsuperscript{\rm 1}\footnotemark[1] \qquad
    Jiaxuan Ren\textsuperscript{\rm 1} \qquad
    Siran Peng\textsuperscript{\rm 2, \rm 3} \qquad
    Liangjian Deng\textsuperscript{\rm 1}\thanks{Corresponding author.}\\
    \textsuperscript{\rm 1}University of Electronic Science and Technology of China \\
    \textsuperscript{\rm 2}Institute of Automation, Chinese Academy of Sciences\\
    \textsuperscript{\rm 3}School of Artificial Intelligence, University of Chinese Academy of Sciences\\
    {\tt\small \{jayhuang,hrchen,jiaxuan.ren\}@std.uestc.edu.cn,liangjian.deng@uestc.edu.cn}
}

\newif\ifreview 
\newif\ifarxiv \newcommand{\arxiv}{\arxivtrue}
\newif\ifcamera 
\newif\ifrebuttal 

\arxiv % \review OR \arxiv OR \cameraready

\pdfoutput=1
\documentclass[10pt,twocolumn,letterpaper]{article}
\ifreview \usepackage[review]{cvpr} \fi
\ifarxiv \usepackage[pagenumbers]{cvpr} \fi
\ifrebuttal \usepackage[rebuttal]{cvpr} \fi
\ifcamera \usepackage{cvpr} \fi

\usepackage{graphicx}	
\usepackage{amsmath}	
\usepackage{amssymb}	
\usepackage{booktabs}
\usepackage{times}
\usepackage{microtype}
\usepackage{epsfig}
\usepackage{caption}
\usepackage{float}
\usepackage{placeins}
\usepackage{color, colortbl}
\usepackage{stfloats}
\usepackage{enumitem}
\usepackage{tabularx}
\usepackage{xstring}
\usepackage{multirow}
\usepackage{xspace}
\usepackage{url}
\usepackage{xcolor}
\usepackage[hang,flushmargin]{footmisc}

\usepackage{pifont}
\usepackage{fancyhdr}
\usepackage[accsupp]{axessibility}

\ifcamera \usepackage[accsupp]{axessibility} \fi

% ...

\ifarxiv  \fi

\newcommand{\R}[1]{{%
    \textbf{%
        \ifstrequal{#1}{1}{\textcolor{red}{R#1}}{%
        \ifstrequal{#1}{2}{\textcolor{blue}{R#1}}{%
        \ifstrequal{#1}{3}{\textcolor{magenta}{R#1}}{%
        \ifstrequal{#1}{4}{\textcolor{teal}{R#1}}{%
                           \textcolor{cyan}{R#1}%
        }}}}%
    }%
}}
\usepackage{xr-hyper}

\makeatletter
\newcommand*{\addFileDependency}[1]{
  \typeout{(#1)}
  \@addtofilelist{#1}
  \IfFileExists{#1}{}{\typeout{No file #1.}}
}

\makeatother
\newcommand*{\myexternaldocument}[1]{
    \externaldocument{#1}
    \addFileDependency{#1.tex}
    \addFileDependency{#1.aux}
}

\definecolor{cvprblue}{rgb}{0.21,0.49,0.74}
\usepackage[pagebackref,breaklinks,colorlinks,allcolors=cvprblue]{hyperref}
\usepackage[capitalize]{cleveref}
\crefname{section}{Sec.}{Secs.}
\crefname{table}{Table}{Tables}
\crefname{figure}{Fig.}{Figs.}

\ifarxiv \crefname{appendix}{App.}{Apps.}
\else \crefname{appendix}{Suppl.}{Suppls.} \fi

\unless\ifarxiv \myexternaldocument{_supplementary} \fi

\begin{document}
%% TITLE
\title{\paperTitle}
\author{\authorBlock}
\maketitle

\ifarxiv
\thispagestyle{fancy}
\fancyhead{}
\fancyhead[C]{\footnotesize This is a pre-print of the original paper accepted at the CVPR Conference on Computer Vision and Pattern Recognition 2025.}
\fi

\begin{abstract}
Currently, deep learning-based methods for remote sensing pansharpening have advanced rapidly. However, many existing methods struggle to fully leverage feature heterogeneity and redundancy, thereby limiting their effectiveness. 
We use the covariance matrix to model the feature heterogeneity and redundancy and propose Correlation-Aware Covariance Weighting (CACW) to adjust them. CACW captures these correlations through the covariance matrix, which is then processed by a nonlinear function to generate weights for adjustment.
Building upon CACW, we introduce a general adaptive dual-level weighting mechanism (ADWM) to address these challenges from two key perspectives, enhancing a wide range of existing deep-learning methods.
First, Intra-Feature Weighting (IFW) evaluates correlations among channels within each feature to reduce redundancy and enhance unique information. 
Second, Cross-Feature Weighting (CFW) adjusts contributions across layers based on inter-layer correlations, refining the final output. 
Extensive experiments demonstrate the superior performance of ADWM compared to recent state-of-the-art (SOTA) methods. Furthermore, we validate the effectiveness of our approach through generality experiments, redundancy visualization, comparison experiments, key variables and complexity analysis, and ablation studies. Our code is available at \url{https://github.com/Jie-1203/ADWM}.
\end{abstract}

\vspace{-0.3cm}
\section{Introduction}
\label{sec:intro}

\begin{figure}[ht]
\centering
\includegraphics[width=1.0\linewidth]{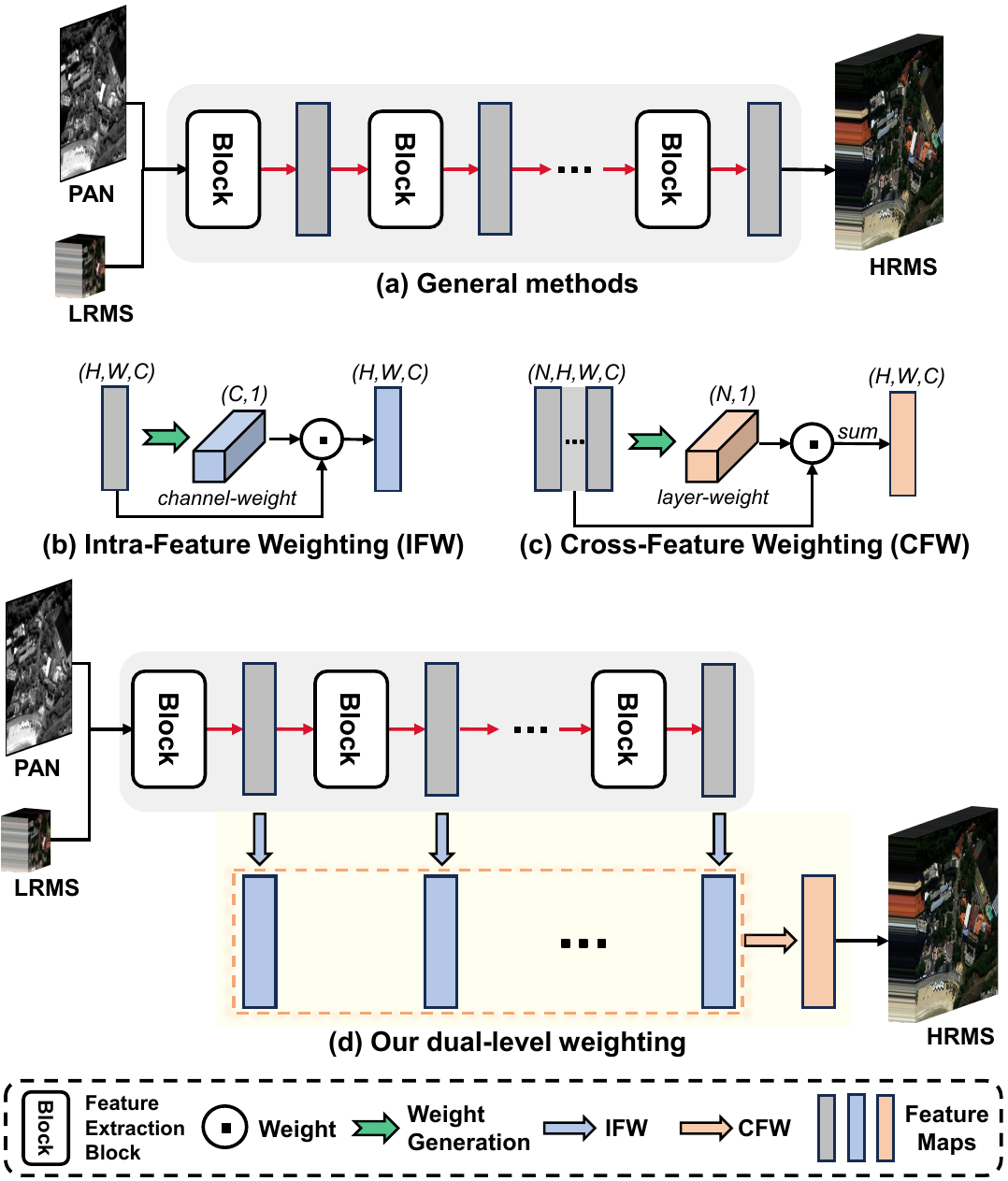}
\caption{
Application of our dual-level weighting mechanism within the existing methods.
(a) General methods. (b) Intra-Feature Weighting (IFW): weighting different channels within a single feature. (c) Cross-Feature Weighting (CFW): weighting features at different depths to obtain the final result. (d) Our dual-level weighting combines both IFW and CFW to fully unlock the potential of the original networks.}
\label{fig:head}
\vspace{-0.7cm}
\end{figure}

The application of high-resolution multispectral (HRMS) images is expanding rapidly, with uses in fields such as object detection~\cite{cheng2016survey,liu2021sraf}, change detection~\cite{asokan2019change, jianya2008review}, unmixing~\cite{bioucas2012hyperspectral}, and classification~\cite{cao2018hyperspectral, cao2020hyperspectral}. However, due to technological limitations, satellites are typically only able to capture low-resolution multispectral (LRMS) images alongside high-resolution panchromatic (PAN) images. In this setup, PAN images provide high spatial resolution, whereas LRMS images offer rich spectral information. To generate HRMS images with both high spatial and spectral resolutions, the pansharpening technique was introduced and has since seen continuous development.

Over recent decades, pansharpening methods have evolved considerably, transitioning from traditional approaches to modern deep learning-based techniques. Traditional methods encompass component substitution (CS) \cite{choiNewAdaptiveComponentSubstitutionBased2011,BDSD-PC}, multi-resolution analysis (MRA) 
\cite{vivoneContrastErrorBasedFusion2014, MTF-GLP-FS}, and variational optimization (VO) \cite{fuVariationalPanSharpeningLocal2019,tianVariationalPansharpeningExploiting2022}. 
With advancements in hardware and software, deep learning-based methods~\cite{ZHUANG2019177Pan-GGF,Yang2020Dual-Injection,dian2018deep,Fu2021DeepMultiscaleDetail} have shown significant promise in addressing pansharpening challenges. 
\begin{figure}[t]
   \centering
   \includegraphics[width=1.0\linewidth]{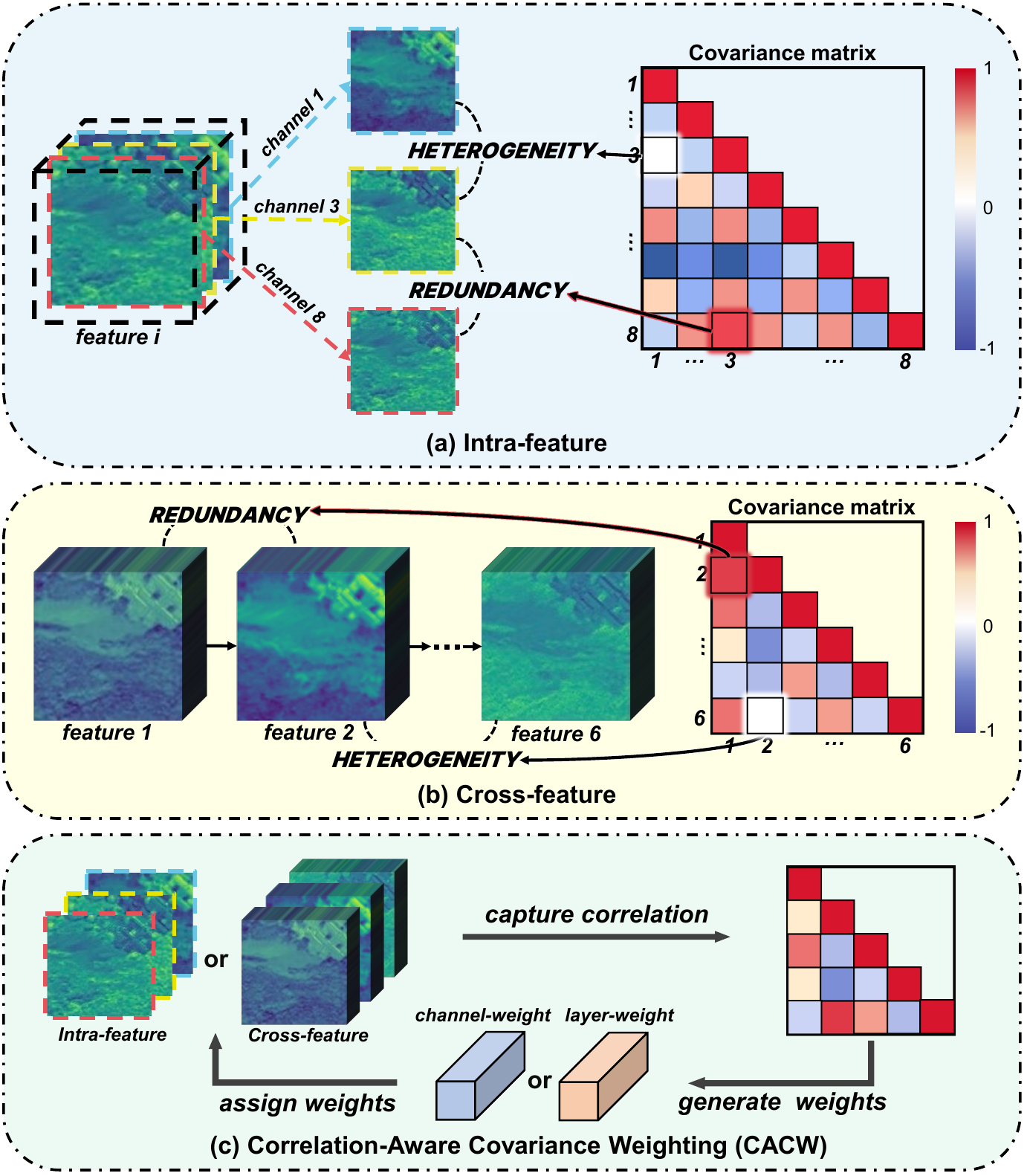}
   \caption{
   Feature heterogeneity and redundancy correspond to the covariance matrix: darker colors indicate stronger correlations and redundancy, while lighter colors suggest weaker correlations and more heterogeneity. (a) Intra-feature: different channels within a feature. (b) Cross-feature: features at different depths. (c) CACW leverages intra- and cross-feature correlations to generate weights and adjust features accordingly.
   }
   \label{fig:a+b}
   \vspace{-0.7cm}
\end{figure}
Recent studies focusing on fusion tasks~\cite{huang2025wavelet, xie2024fusionmamba,li2018CoupledSparseTensorFactorization} highlight the benefits of sequential feature extraction to achieve effective data fusion.
This keeps the continuity with the tradition of using sequential feature extraction as shown in \cref{fig:head} (a) for achieving improved fusion in pansharpening.

However, the above methods often overlook a crucial characteristic across networks: feature heterogeneity and redundancy, which occur in two dimensions as shown in \cref{fig:a+b}. In the intra-feature dimension, PAN and LRMS images contain redundant and heterogeneous information, with multiple bands in LRMS images exhibiting similar redundancy~\cite{ghassemian2016review}. In the cross-feature dimension, shallow features capture low-level details like edges and textures, while deeper features become abstract, encoding semantic information~\cite{Zeiler2014Visualizing}. These features vary by depth but also exhibit redundancy. Overall, previous methods fail to comprehensively address these issues, limiting refined fusion.

% para4 
To overcome the limitations of previous methods, we propose an Adaptive Dual-level Weighting Mechanism (ADWM). 
Our method can be seamlessly plugged into the network involving sequential feature processing to fully unleash the potential of the original network, see \cref{fig:head}.
Specifically, our first level of weighting, termed Intra-Feature Weighting (IFW), focuses on optimizing the internal structure of intermediate features through adaptive weighting. 
This step emphasizes the importance of each channel individually, allowing for a refined adjustment that accounts for the feature heterogeneity and redundancy channels themselves.
Following this, the second level of weighting, Cross-Feature Weighting (CFW), dynamically adjusts the contribution of features from different layers based on inter-layer correlations. This approach allows features at various depths to contribute adaptively to the final output, ensuring a balanced integration of deep-shallow features across the network.
Unlike DenseNet~\cite{Huang_Liu_Van_2017}, which directly concatenates feature maps from different layers, CFW adjusts each layer’s influence based on its relevance to the final output. CFW balances shallow and deep layer contributions, enhancing representation precision and fidelity. 

To achieve the above process, we need a weighting method that can finely capture relationships between features.
SENet~\cite{Hu2013senet} uses global pooling for channel-level weights, applying high-level adaptive weighting but oversimplifies feature compression, which leads to a loss of inter-channel relationships. SRANet~\cite{ling2019Sra} employs a self-residual attention mechanism for dynamic channel weights but lacks effective redundancy compression, leading to suboptimal resource use.
\textit{Different from previous benchmark attention-based methods, such as channel attention~\cite{wang2020eca,ling2019Sra}, spatial attention~\cite{Hu2013senet}, etc., our method leverages the correlations within the covariance matrix, effectively capturing feature heterogeneity and redundancy, forming the proposed Correlation-Aware Covariance Weighting (CACW).}
The use of the covariance matrix is inspired by Principal Component Analysis (PCA)~\cite{abdi2010principal}, which utilizes the covariance matrix to capture information correlation for dimensionality reduction. 
Specifically, as shown in \cref{fig:a+b} (c), we first compute a covariance matrix to capture inter- or cross-feature correlations, reflecting feature heterogeneity and redundancy. 
Subsequently, we further process this covariance matrix through a nonlinear transformation enabling the generation of weights for feature adjustment. 
By leveraging these weights, the model identifies the importance of different channels or layers, focusing on essential components while effectively reducing redundancy.

To sum up, the contributions of this work are as follows:
\begin{enumerate} 
\item We use covariance to measure feature heterogeneity and redundancy and propose CACW, which utilizes the correlations within the covariance matrix to adaptively generate weights. This weighting strategy can also be extended to applications beyond pansharpening.
\item 
We propose an adaptive dual-level weighting mechanism (ADWM). 
This mechanism uses IFW to adjust the importance of different channels in the intra-feature dimension and applies CFW to adaptively modulate the contribution of shallow and deep features to the final result in the cross-feature dimension.
\item Extensive experiments verify that the ADWM module can be seamlessly integrated into various existing networks, enhancing their performance and achieving state-of-the-art results in a plug-and-play manner.
\end{enumerate}

\begin{figure*}[ht]
  \centering
  \includegraphics[width=0.9\linewidth]{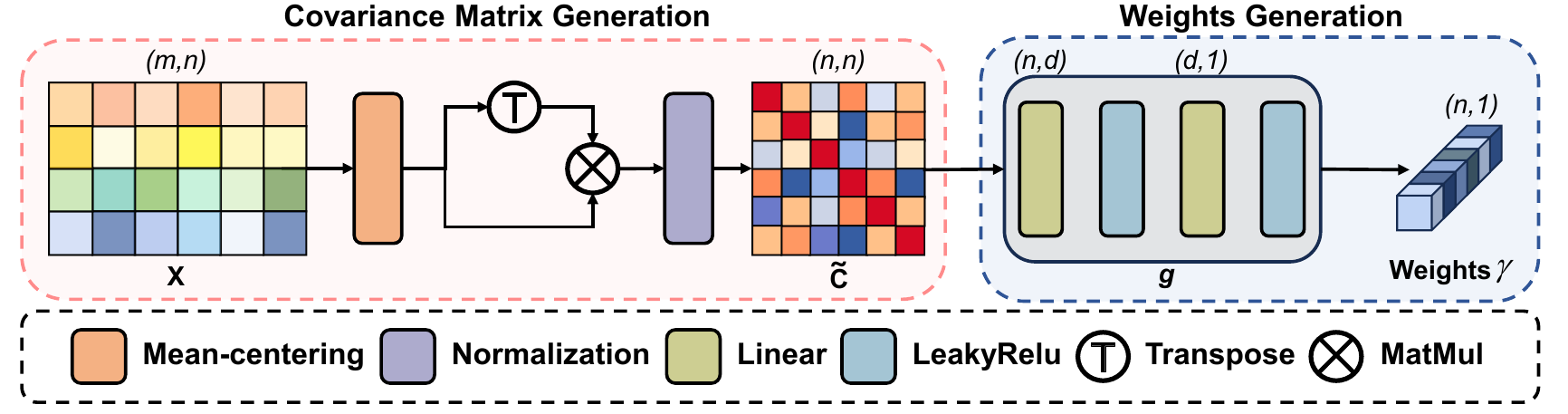}
  \vspace{-0.4cm}
  \caption{
The CACW structure is illustrated. First, we compute the covariance matrix \( \tilde{C} \) based on the correlations among the \( n \) columns of \( X \). Then, this covariance matrix is passed through a nonlinear function \( g \), which generates the resulting weights.}
  \label{fig:weighting}
  \vspace{-0.6cm}
\end{figure*}
\vspace{-2pt}

\section{Method}
\label{sec:method}

In this section, we first introduce CACW, followed by an overview of ADWM, a detailed explanation of the proposed IFW and CFW, and a supplement on the plug-and-play integration of ADWM.

\vspace{-2pt}

\begin{figure*}[t]
  \centering
  \includegraphics[width=1.0\linewidth]{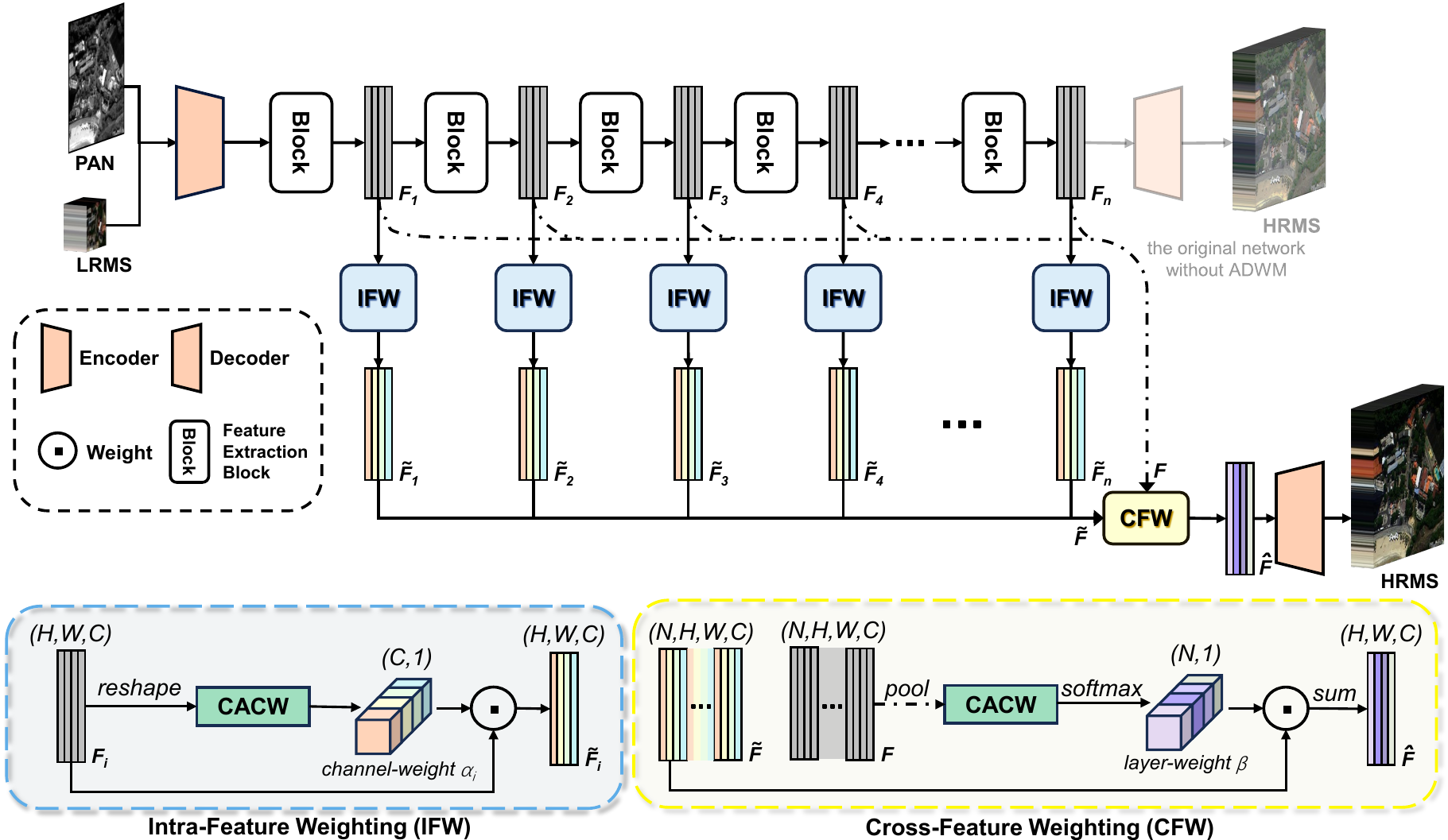}
  \vspace{-0.4cm}
  \caption{
The overall workflow of ADWM is comprised of two sub-modules: Intra-Feature Weighting (IFW) and Cross-Feature Weighting (CFW). In IFW, each original feature \( F_i \) is adjusted to \( \tilde{F}_i \) based on its internal correlations. In CFW, weights are generated based on the correlations among \( F_i \) features, dynamically adjusting each \( \tilde{F}_i \)'s contribution to the final output.
}
  \label{fig:overall}
  \vspace{-0.5cm}
\end{figure*}
\vspace{3pt}

\subsection{Correlation-Aware Covariance Weighting}
\label{sec:f}
In this section, we will introduce the design of CACW. \cref{fig:weighting} illustrates the overall process of CACW.

\vspace{3pt}
\noindent{\bf Background of PCA}~\cite{abdi2010principal}. 
PCA is a common technique for dimensionality reduction.
Let \( X \in \mathbb{R}^{m \times n} \) represent the input observation matrix, where \( m \) denotes the number of samples and \( n \) denotes the number of features to be explored. The covariance matrix \( C \in \mathbb{R}^{n \times n} \) can be calculated as follows:
\begin{equation}
\vspace{-3pt}
\label{equ:C}
C = \frac{1}{m-1} (X - \bar{X})^T (X - \bar{X}),
\end{equation}
where \( \bar{X} \) is the mean of each feature in \( X \). 
Next, we perform eigenvalue decomposition and select the top \( k \) eigenvectors \( \mathbf{v}_i \) with the largest eigenvalues \( \lambda_i \) to form matrix \(P = [\mathbf{v}_1, \mathbf{v}_2, \ldots, \mathbf{v}_k] \). 
The process of generating \( \mathbf{v}_i \) can be formulated as follows:
\vspace{-2pt}
\begin{equation}
%\vspace{-2pt}
\label{equ:eigen}
C \mathbf{v}_i = \lambda_i \mathbf{v}_i.
\end{equation}
By projecting \( X \) onto this subspace using \( P \), we obtain a reduced-dimensional representation \( Y \). The process of projecting is as follows:
\vspace{-2pt}
\begin{equation}
\label{equ:projecting}
Y = P^T X.
\end{equation}

\vspace{-2pt}
\noindent{\bf CACW}. 
The purpose of CACW is to generate weights for feature selection through adaptive adjustment, rather than dimensionality reduction via projection.
In PCA, eigenvectors \( \mathbf{v}_i \) represent the main directions of variation, while in CACW, the weight generation process establishes a basis to highlight important features and suppress redundancy. We first compute the covariance matrix \( C \) using \cref{equ:C}, capturing correlations that reflect feature heterogeneity and redundancy.
To mitigate the influence of absolute magnitudes, we normalize \( C \) to obtain \( \tilde{C} \). The process of normalization is as follows:
\begin{equation}
\tilde{C}_{ij} = \frac{C_{ij}}{\|\bar{X_i}\| \|\bar{X_j}\|},
\vspace{-1pt}
\end{equation}
where \( \|\bar{X_i}\| \) and \( \|\bar{X_j}    \| \) are the norms of the respective feature vectors and each element \( \tilde{C}_{ij} \) represents the normalized similarity between features \( i \) and \( j \).
As shown in \cref{equ:eigen}, PCA's eigenvalue decomposition relies on linear feature decomposition, which is not data-driven and cannot adaptively highlight important features while suppressing redundancy. To address this, we achieve linear feature decomposition through a neural network. Specifically, we use a multi-layer perceptron (MLP) \( g \) to nonlinearly map the covariance matrix \( \tilde{C} \) to obtain the weight vector \( \gamma \in \mathbb{R}^{n \times 1} \).
The process of generating weights is as follows:
\begin{equation}
\gamma = g(\tilde{C}),
\end{equation}
where \( \gamma \) can represent either channel-weight in IFW or layer-weight in CFW.

\noindent{\bf Difference with attention}. 
CACW stems from the covariance-based observation in \cref{fig:a+b} and the motivation to reduce feature redundancy and enhance heterogeneity, rather than being inspired by attention. Due to the difference between inspiration and motivation, its operation also fundamentally differs from attention.
The scaling normalizes the covariance matrix to a correlation matrix, guiding the MLP to capture correlations, while attention’s scaling mainly addresses gradient stability.
CACW's MLP is placed after the covariance matrix for PCA-like eigenvalue decomposition in a data-driven way, while attention's MLP comes before QKV to extract nonlinear features, ignoring data correlations.

\vspace{-2pt}
\subsection{Overview of ADWM}
\vspace{-1pt}
We denote PAN as $P\in\mathbb{R}^{H\times W}$, LRMS as $L \in \mathbb{R}^{\frac{H}{4} \times \frac{W}{4} \times c}$, and HRMS as $H \in \mathbb{R}^{H\times W\times c}$.
In many methods, \( P \) and \( L \) are processed through an encoder and then fed into a sequential feature extraction block, such as Resblock~\cite{He_Zhang_Ren_Sun_2016}, generating multiple intermediate features \( F_i \in \mathbb{R}^{H \times W \times C} \), where \( i \) represents the \( i \)-th layer. 
Our ADWM is applied to adjust these features, maintaining generality and flexibility by integrating directly into existing network architectures without requiring modifications to the original structure.
Firstly, each feature \( F_i \) is adaptively weighted based on its own correlation, resulting in \( \tilde{F}_i \in \mathbb{R}^{H \times W \times C} \).
Then, we collect \( F_i \) and \( \tilde{F}_i, \) \( i = 1, 2, \dots, n \) to obtain \( F \) and \( \tilde{F} \in \mathbb{R}^{N \times H \times W \times C} \).
We generate weights based on \( F \) and apply them to \( \tilde{F} \), resulting in the adjusted \( \hat{F} \). The dual-level weighting process can be formulated as follows:
\vspace{-3pt}
\begin{equation}
\tilde{F_i} = \text{IFW}(F_i), 
\end{equation}
\vspace{-5pt}
\begin{equation}
F = [F_1, \dots, F_n], \quad \tilde{F} = [\tilde{F}_1, \dots, \tilde{F}_n],
\end{equation}
\vspace{-5pt}
\begin{equation}
\hat{F} = \text{CFW}(F, \tilde{F}),
\end{equation}
where 
IFW and CFW represent the intra-feature and cross-feature weighting processes, respectively, which are detailed in \cref{sec:IFW} and \cref{sec:CFW}. \( \hat{F} \) is then processed by a decoder to obtain the final output \( H \).

\subsection{Intra-Feature Weighting}
\label{sec:IFW}
\noindent\textbf{The details of IFW:}
First, we reshape \( F_i \) to obtain \( F_i^R \in \mathbb{R}^{HW \times C} \), focusing on exploring the correlations between channels by treating the \( HW \) spatial pixels as our sample space. Second, to dynamically adjust each feature \( F_i \), we calculate the channel-weight \( \alpha_i \in \mathbb{R}^{C \times 1} \), which addresses the heterogeneity and redundancy within the feature by assigning individualized importance to each channel based on its unique contribution. The process of generating channel-weight \( \alpha_i \) is as follows:
\begin{equation}
F_i^R = \text{Reshape}(F_i),
%\vspace{-5pt}
\end{equation}
\vspace{-5pt}
\begin{equation}
\alpha_i = f(F_i^R),
\end{equation}
where \( f(\cdot) \) denotes the CACW.
The channel-weight \( \alpha_i \) is then applied to \( F_i \), adjusting the relative importance of each channel to produce the adjusted feature \( \tilde{F}_i \). The weighting process is as follows:
\begin{equation}
\tilde{F_i} = F_i \odot \alpha_i,
\vspace{-3pt}
\end{equation}
where \( \odot \) denotes element-wise production.
Unlike the PCA's global orthogonal projection in \cref{equ:projecting}, IFW independently scales each feature dimension through the generated weights. This can be seen as a simplified form of projection that preserves the original feature basis while optimizing the distribution of information.

\begin{figure}[t]
   \centering
   \includegraphics[width=1.0\linewidth]{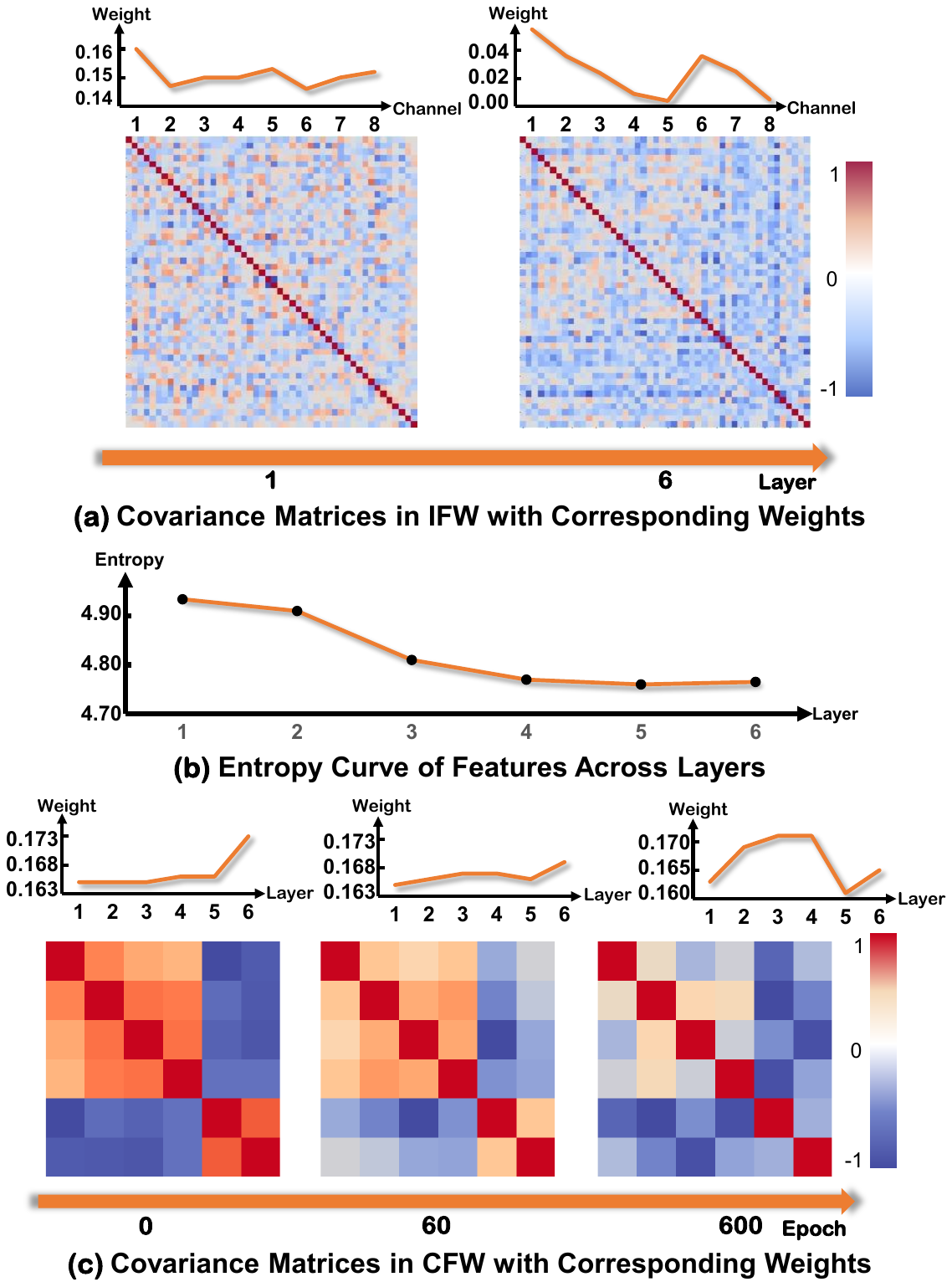}
   \vspace{-0.6cm}
   \caption{Visualization of covariance matrices, weights in IFW and CFW. (a) Channels that are multiples of six selected for clarity. (b) Lower entropy indicates higher feature redundancy.}
   \label{fig:visual_new}
   \vspace{-0.7cm}
\end{figure}

% \noindent{\bf CACW}. 
\noindent\textbf{The impact of IFW:}
As shown in \cref{fig:visual_new} (a), shallower layers have lighter matrices, reflecting greater information diversity, while deeper layers darken, indicating high redundancy, as confirmed by the decreasing entropy trend in \cref{fig:visual_new} (b). IFW adjusts weights accordingly, producing varying distributions across layers. In shallow layers, weights are uniform (0.141–0.155), while in deeper layers, they vary more (0.004–0.055), emphasizing key channels and suppressing others, helping the model focus on critical structures.

\subsection{Cross-Feature Weighting}
\label{sec:CFW}
\noindent\textbf{The details of CFW}:
The goal of CFW is to use adaptive weighting to fully leverage all intermediate features, effectively addressing heterogeneity and redundancy across layers.
First, we apply spatial average pooling to adjust the shape of \( F \), resulting in a representation \( F^P \in \mathbb{R}^{C\times N} \), which enables us to explore the correlations between the \( N \) layers by treating the \( C \) channels as our sample space.
Second, to dynamically adjust the contribution of different layer features to the final result, we calculate the layer-weight \( \beta \in \mathbb{R}^{N \times 1} \). The process of generating channel-weight \( \beta \) can be formulated as follows:
\vspace{-4pt}
\begin{equation}
\vspace{-1pt}
F^P = \frac{1}{H \times W} \sum_{i=1}^H \sum_{j=1}^W F_{ij},
\end{equation}
\vspace{-5pt}
\begin{equation}
 \beta = f(F^P),
\end{equation}
where the reshape operation flattens the width and height of the original input and \( f(\cdot) \) denotes the CACW.
Unlike IFW, CFW produces a single output feature \( \hat{F} \) synthesized from \( N \) intermediate features. 
First, we use softmax on \( \beta \) to normalize layer weights, balancing depth contributions. Then, these weights modulate each intermediate feature \( \tilde{F} \) reducing redundancy by giving less weight to repetitive layers. A final summation produces the integrated feature \( \hat{F} \in \mathbb{R}^{H \times W \times C} \). 
The weighting process can be formulated as follows:
\vspace{-5pt}
\begin{equation}
\label{equ:12}
\vspace{-5pt}
\begin{aligned}
\hat{F} = \sum_{k=1}^N \left( \tilde{F} \odot \textrm{softmax}(\beta) \right),
\end{aligned}
\end{equation}
where \( \odot \) denotes element-wise production, and \( \sum_{k=1}^N \) represents the summation across the \( N \)-dimension. The pointwise weighting and summation process in \cref{equ:12} can also be rewritten in a matrix multiplication form as follows:
\begin{equation}
\label{equ:2}
\hat{F} = (\textrm{softmax}(\beta))^T \tilde{F}.
\end{equation}
This corresponds to \cref{equ:projecting} in form. However, unlike PCA's global dimensionality reduction using a fixed orthogonal basis, CFW dynamically learns and applies task-specific weights \( \beta \).

\begin{figure}[t]
   \centering
   \includegraphics[width=1.0\linewidth]{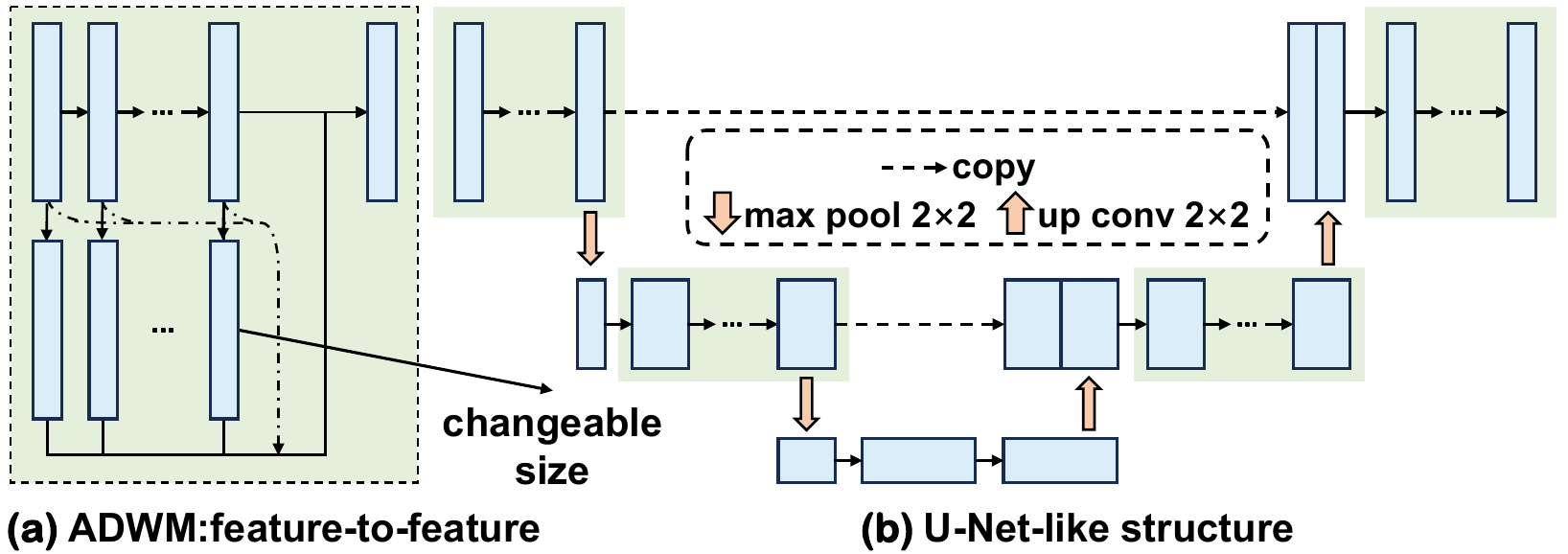}
   \caption{(a) ADWM can be seen as a feature-to-feature method. (b) The way it is embedded into more complex networks.}
   \label{fig:Unet}
   \vspace{-0.7cm}
\end{figure}

\noindent\textbf{The impact of CFW}:
As shown in \cref{fig:visual_new} (c), the CFW covariance matrix evolves during training, transitioning from deep red and blue regions to an overall lighter color, indicating reduced redundancy and increased feature diversity.
As training progresses, the layer weights generated by CFW adjust gradually, reflecting the model’s adaptive tuning of each layer's impact. These shifts reflect the model’s refinement to align layer contributions with task demands, enhancing its ability to leverage diverse features and optimize performance throughout learning.

\begin{figure*}[t]
  \centering
  \includegraphics[width=1.0\linewidth]{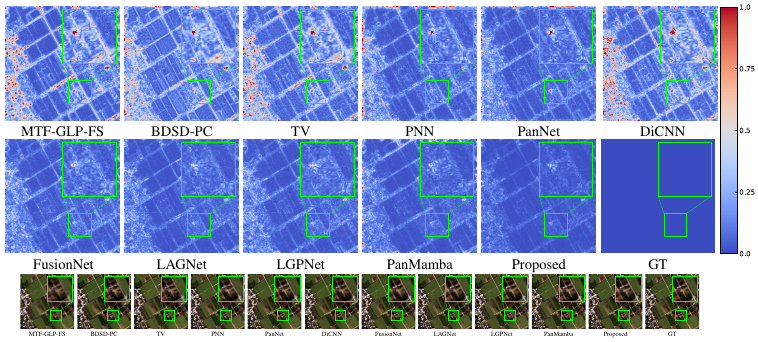}
  \captionsetup{justification=raggedright, singlelinecheck=false}
  \vspace{-0.6cm}
  \caption{The residuals (Top) and visual results (bottom) of all compared approaches on the GF2 reduced-resolution dataset.}
  \label{fig:WV3reduce}
  \vspace{-0.2cm}
\end{figure*}
\begin{table*}[ht]
\centering
\caption{Comparisons on WV3, QB, and GF2 reduced-resolution datasets, each with 20 samples. Best:\textbf{bold} , and second-best: \underline{underline}.}
\vspace{-0.1cm}
\begin{tabular}{cc@{\hskip 3pt}c@{\hskip 0.005in}c@{\hskip 0.005in}cc@{\hskip 3pt}c@{\hskip 0.005in}c@{\hskip 0.005in}cc@{\hskip 3pt}c@{\hskip 0.005in}c@{\hskip 0.005in}c}
\hline
\vspace{-3pt}
\multirow{2}{*}{\textbf{Methods}} & \multicolumn{4}{c}{\textbf{ WV3 }} & \multicolumn{4}{c}{\textbf{QB}} & \multicolumn{4}{c}{\textbf{GF2}}\\ 
\cmidrule(lr){2-5}  \cmidrule(lr){6-9}  \cmidrule(lr){10-13}  
 & \textbf{PSNR$\uparrow$} & \textbf{SAM$\downarrow$} & \textbf{ERGAS$\downarrow$} & \textbf{Q8$\uparrow$}& \textbf{PSNR$\uparrow$} & \textbf{SAM$\downarrow$} & \textbf{ERGAS$\downarrow$} & \textbf{Q4$\uparrow$} & \textbf{PSNR$\uparrow$} & \textbf{SAM$\downarrow$} & \textbf{ERGAS$\downarrow$} &\textbf{Q4$\uparrow$} 
\\ \hline
MTF-GLP-FS~\cite{MTF-GLP-FS} &32.963 & 5.316 & 4.700 & 0.833  & 32.709 & 7.792 & 7.373 &0.835 & 35.540 & 1.655 & 1.589 & 0.897
\\  
BDSD-PC~\cite{BDSD-PC} &32.970 & 5.428 & 4.697 & 0.829 & 32.550 & 8.085 & 7.513 &0.831 & 35.180 & 1.681 & 1.667 & 0.892 
\\ 
TV~\cite{TV} & 32.381 & 5.692 & 4.855 & 0.795& 32.136 & 7.510 & 7.690 &0.821 & 35.237 & 1.911 & 1.737 & 0.907 
\\ \hline

PNN~\cite{PNN} & 37.313 & 3.677 & 2.681 & 0.893  & 36.942 & 5.181 & 4.468 &0.918 & 39.071 & 1.048 & 1.057 & 0.960 
\\ 
PanNet~\cite{Yang2017PanNet} &37.346 & 3.613 & 2.664 & 0.891  & 34.678 & 5.767 & 5.859 &0.885& 40.243 & 0.997 & 0.919 & 0.967
\\ 
DiCNN~\cite{hePansharpeningDetailInjection2019} & 37.390 & 3.592 & 2.672 & 0.900 & 35.781 & 5.367 & 5.133 & 0.904 & 38.906 & 1.053 & 1.081 & 0.959 \\ 
FusionNet~\cite{FusionNet} & 38.047 & 3.324 & 2.465 & 0.904 & 37.540 & 4.904 & 4.156 &0.925 & 39.639 & 0.974 & 0.988 & 0.964 
\\ 
LAGNet~\cite{LAGConv}& 38.592& 3.103& 2.291& 0.910& 38.209& 4.534& 3.812&\underline{0.934}& 42.735& 0.786& 0.687& 0.980
\\ 
LGPNet~\cite{zhao2023lgpconv}& 38.147& 3.270& 2.422& 0.902& 36.443& 4.954& 4.777& 0.915& 41.843& 0.845& 0.765&  0.976
\\ 
PanMamba~\cite{He2024PanMambaEP} & \underline{39.012}& \textbf{2.913}& \underline{2.184}& \underline{0.920} & 37.356 & 4.625 & 4.277 &0.929 & 42.907 & 0.743& 0.684 & 0.982
\\ 
% HMPNet&0& 38.765& 3.063& 2.230& 0.916& \textbf{39.227}& 4.617& \textbf{3.404}& 0.936& \textbf{45.926}& 0.782& \textbf{0.537}& \underline{0.985}\\
%  ZS-Pan~\cite{caoif2024} & 39.001       & 2.938& 2.192& 0.918
%  &\underline{38.211}& \underline{4.512}&  \underline{3.807}&0.930& 
%  \underline{43.286}& \underline{0.715}& \underline{0.634}&\underline{0.983}
% \\ 
 
 \textbf{Proposed} & \textbf{39.170}& \textbf{2.913}& \textbf{2.145}& \textbf{0.921}& \textbf{38.466}& \textbf{4.450}& \textbf{3.705}&\textbf{0.937}& \textbf{43.884}& \textbf{0.672}& \textbf{0.597}& \textbf{0.985} \\ \hline
\end{tabular}
%\caption{Comparisons on WV3, QB, and GF2 reduced-resolution datasets, each with 20 samples, respectively. Best: bold, and second-best: underline.}
\label{tab:all_reduce}
\vspace{-0.4cm}
\end{table*}

\subsection{Flexible Plug-and-Play Integration}
\vspace{-5pt}
In addition to being integrated with the entire network in the way shown in \cref{fig:overall}, ADWM can also be flexibly combined with more complex networks.
As shown in \cref{fig:Unet} (a), it can be seen as a feature-to-feature method, deriving the next feature from a series of sequential features.
Especially, as shown in \cref{fig:Unet} (b), complex networks like U-Net often consist of parts with similar feature sizes and semantics, and an independent ADWM is applied to each part.
As shown in \cref{table:general}, it still improves performance even when applied to a more complex network.
\begin{figure*}[ht]
  \centering
  \includegraphics[width=1.0\linewidth]{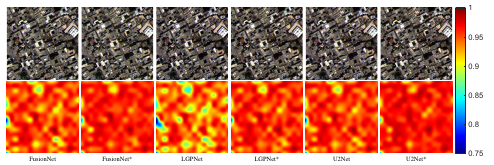}
  \captionsetup{justification=raggedright, singlelinecheck=false}
  \vspace{-0.5cm}
  \caption{
The visual results (Top) and HQNR maps (Bottom) of all evaluated general methods on the WV3 full-resolution dataset.
  }
  \label{fig:hot}
  \vspace{-0.1cm}
\end{figure*}
\begin{table*}[ht]
\centering
\caption{Comparisons on WV3, QB, and GF2 datasets with 20 full-resolution samples, respectively. Methods marked with * represent the corresponding method enhanced with our ADWM module without any further changes. The best results in each column are \textbf{bolded}.}
\vspace{-0.1cm}
	\begin{tabular}{ccc@{\hskip 7pt}c@{\hskip 5pt}cc@{\hskip7pt}c@{\hskip5pt}cc@{\hskip7pt}c@{\hskip 5pt}c}
		\hline
        \vspace{-3pt}
		\multirow{2}{*}{\textbf{Method}} & \multirow{2}{*}{\textbf{Params}} & \multicolumn{3}{c}{\textbf{WV3}}                                              & \multicolumn{3}{c}{\textbf{QB}}                                               & \multicolumn{3}{c}{\textbf{GF2}}                                 \\ 
        \cmidrule(lr){3-5}  \cmidrule(lr){6-8}  \cmidrule(lr){9-11}  
		&                & \textit{\textbf{D$_\lambda \downarrow$}} & \textit{\textbf{D$_s \downarrow$}}  & \textbf{HQNR↑} & \textit{\textbf{D$_\lambda \downarrow$}}  & \textit{\textbf{D$_s \downarrow$}} & \textbf{HQNR↑} & \textit{\textbf{D$_\lambda \downarrow$}} & \textit{\textbf{D$_s \downarrow$}} & \textbf{HQNR↑} \\ \hline
		\textbf{FusionNet}~\cite{FusionNet}               &                                 78.6K& 0.024                  & 0.037          &0.940          & \textbf{0.057}& 0.052                  & 0.894         & 0.035                  & 0.101                  & 0.867          \\
		\textbf{FusionNet*}              &                                 85.5K& \textbf{0.022}& \textbf{0.033}& \textbf{0.946} & 0.066& \textbf{0.038}& \textbf{0.899} & \textbf{0.034}& \textbf{0.094}& \textbf{0.875}\\ \hline
        \textbf{LGPNet}~\cite{zhao2023lgpconv}                  &                                 27.0K& 0.022& 0.039& 0.940          & 0.093          & 0.061& 0.870          & 0.030& 0.081& 0.892\\
		\textbf{LGPNet*}&                                 33.4K& \textbf{0.020}& \textbf{0.032}&\textbf{0.950} & \textbf{0.068}& \textbf{0.041}&\textbf{0.894} & \textbf{0.027}& \textbf{0.068}& \textbf{0.908}\\ \hline
        		\textbf{U2Net}~\cite{U2Net}                   &                                 757.5K& 0.020& 0.028                  & 0.952      & \textbf{0.052}          & 0.037                  & 0.913        &  0.024                  & 0.051          & 0.927            
\\
		\textbf{U2Net*}                  &                                 781.4K& \textbf{0.019}& \textbf{0.026}& \textbf{0.955} &0.054& \textbf{0.029}& \textbf{0.919}& \textbf{0.021}& \textbf{0.049}& \textbf{0.931}\\ \hline
	\end{tabular}
%\caption{Comparisons on WV3, QB, and GF2 datasets with 20 full-resolution samples, respectively. Methods marked with * represent the corresponding method enhanced with our ADWM module without any further changes. Best results in each column are bolded.}
\label{table:general}
\vspace{-0.3cm}
\end{table*}

\section{Experiments}
\label{sec:experiments}

\subsection{Datasets, Metrics, and Training Details}
In our experiments, we employ three datasets derived from satellite imagery captured by WorldView-3 (WV3), QuickBird (QB), and GaoFen-2 (GF2), constructed in accordance with Wald's protocol~\cite{waldFusionSatelliteImages1997}. The datasets and associated data processing techniques are obtained from the PanCollection repository\footnote{\url{https://github.com/liangjiandeng/PanCollection}}~\cite{dengMachineLearningPansharpening2022}.
We use well-established evaluation metrics to assess our method. For the reduced-resolution dataset, we employ SAM~\cite{boardman1993automating}, ERGAS~\cite{wald2002data}, Q4/Q8 ~\cite{garzelliHypercomplexQualityAssessment2009}, and PSNR. For the full-resolution dataset, D$_s$, D$_\lambda$, and HQNR~\cite{arienzo2022full} are used as evaluation metrics. Among them, HQNR is derived from \(D_s\) and \(D_\lambda\), providing a comprehensive assessment of image quality.
In addition, we train our model in \cref{tab:all_reduce} using the $\ell_1$ loss function and the Adam optimizer \cite{Adam} with a batch size of 64. The training details of our method in \cref{table:general} remain consistent with those in the original paper. All experiments are conducted on an NVIDIA GeForce GTX 3090 GPU.
\textit{More details can be found in supplementary materials.}

\subsection{Comparison with SOTA methods}
We evaluate our method against nine competitive approaches, including 1) three classical methods: MTF-GLP-FS~\cite{MTF-GLP-FS}, BDSD-PC~\cite{BDSD-PC}, and TV~\cite{TV}; 2) seven deep learning-based methods PNN~\cite{PNN}, PanNet~\cite{Yang2017PanNet}, DiCNN~\cite{hePansharpeningDetailInjection2019}, FusionNet~\cite{FusionNet}, LAGNet~\cite{LAGConv}, LGPNet~\cite{zhao2023lgpconv}, and PanMamba~\cite{He2024PanMambaEP}. We incorporated ADWM into the classic LAGNet as our proposed method.
\cref{tab:all_reduce} presents a comprehensive comparison of our method with state-of-the-art approaches across three datasets.
Notably, our method achieves this high level of performance across all metrics.
Specifically, our method achieves a PSNR improvement of 0.158dB,
0.255dB, and 0.598dB on the WV3, QB, and GF2 datasets, respectively, compared to the second-best results.
These improvements highlight the clear advantages of our method,
confirming its competitiveness in the field.
\cref{fig:WV3reduce} provides qualitative assessment results for GF2 datasets alongside the respective ground truth (GT). 
By comparing the mean squared error (MSE) residuals between the pansharpened results and the ground truth, it is clear that our residual maps are the darkest, suggesting that our method achieves the highest fidelity among the evaluated methods.

\vspace{-2pt}
\subsection{Generality Experiment}
\vspace{-1pt}
Our ADWM serves as a plug-and-play mechanism, allowing it to be easily integrated into various existing frameworks without requiring extensive modifications.
To further demonstrate the generality of our method, we integrated three different approaches into our ADWM framework, including FusionNet~\cite{FusionNet}, LGPNet~\cite{zhao2023lgpconv}, and U2Net~\cite{U2Net}.
As shown in \cref{table:general}, our method improves performance for each approach across three datasets, with only a negligible increase in parameter size. 
In terms of visual quality shown in \cref{fig:hot}, as depicted in the second row, the redder areas indicate better performance, while the bluer areas indicate poorer performance. After integrating ADWM into all the methods, they all show larger and deeper red areas, indicating better performance.
These results validate our method's significant potential for practical applications.

\vspace{-2pt}
\subsection{Redundancy Visualization}
\begin{figure}[t]
   \centering
   \includegraphics[width=1.0\linewidth]{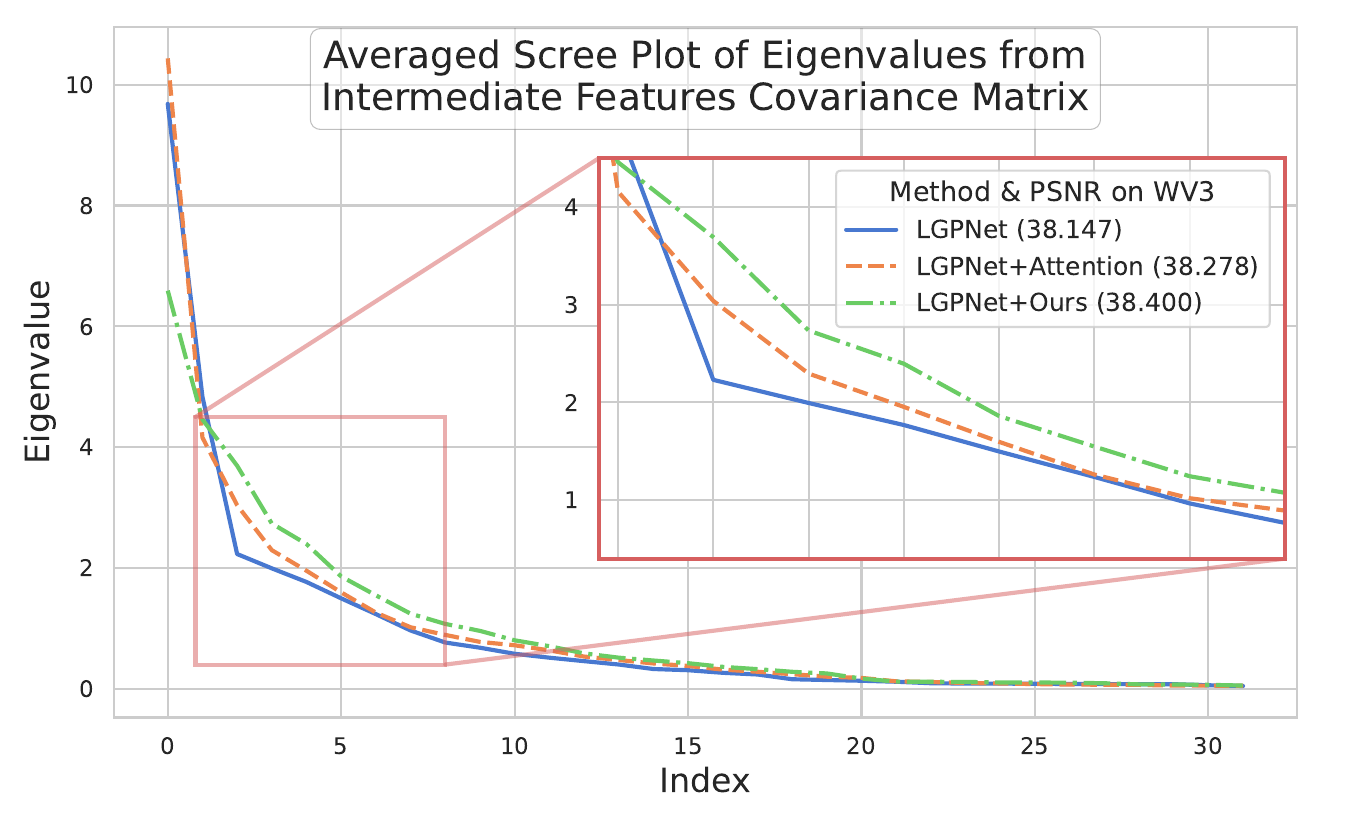}
   \vspace{-0.85cm}
   \caption{Scree Plot to illustrate the differences in redundancy. The smoother the curve, the lower the redundancy.}
   \label{fig:redundancy}
   \label{fig:visual}
   \vspace{-0.3cm}
\end{figure}

\vspace{-1pt}
To demonstrate that our method improves by our motivation to reduce feature redundancy and enhance heterogeneity, we visualized the results using LGPNet~\cite{zhao2023lgpconv} on the WV3 reduced-resolution dataset.
We applied SVD to the covariance matrix of each intermediate feature, sorted the eigenvalues to obtain the corresponding scree plot~\cite{scree_plot}, and averaged all the intermediate features to generate \cref{fig:redundancy}.  
Compared to the original network and self-attention-based weighting, our method yields the smoothest curve, indicating a more balanced distribution of variance across multiple dimensions. This suggests that information is more evenly spread rather than being concentrated in a few dominant components, leading to reduced redundancy. This reduction leads to higher performance metrics, such as PSNR.

\subsection{Evaluation and Analysis of CACW}
\vspace{-1pt}
\noindent\textbf{Comparison of Weighting Approaches}:
This section shows the comparison results of CACW with other commonly used approaches for generating weights.
In the first row of \cref{table:diff_weight}, PCA selects the top half eigenvectors corresponding to the largest eigenvalues, which are then processed by an MLP to generate the weight. CACW outperforms PCA, showing that a neural network with nonlinear capabilities creates a more effective feature importance representation.
In the second row, Pool denotes global average pooling, compressing each channel to a scalar and generating weights through an MLP. CACW surpasses Pool with fewer parameters by preserving inter-channel correlations via a covariance matrix, addressing the information loss and dependency limitations of pooling.
In the third row, Attention uses self-attention to compute a relevance matrix before generating weights through an MLP. With fewer parameters, CACW outperforms Attention, avoiding the complexity and noise of attention mechanisms.

\begin{table}[t]
\centering
\caption{Comparison of different weight generation approaches.
%, with parameters calculated only for the weighting component.
}
\vspace{-0.1cm}
\setlength{\tabcolsep}{4pt} % 默认值为6pt，调整为4pt以缩短列间距
\centering
\begin{tabular}{cccccc}
\hline
\vspace{-3pt}
\multirow{2}{*}{\textbf{Method}} & \multirow{2}{*}{\textbf{Params}} & \multicolumn{4}{c}{\textbf{WV3}}                               \\
\cmidrule(lr){3-6}
                                 &                                  & \textbf{PSNR↑} & \textbf{SAM↓} & \textbf{ERGAS↓} & \textbf{Q8↑} \\ \hline
\textbf{PCA}         &                                  11.2K&               38.612&               3.076&                 2.273&              0.913\\
\textbf{Pool}                &                                  39.2K& 39.053& 2.915& 2.183& 0.920\\
\textbf{Attention}         &                                  42.5K&               38.875&               3.012&                 2.237&              0.915\\
\textbf{CACW}        &                                  26.2K&               \textbf{39.170}&               \textbf{2.913}&                 \textbf{2.145}&            \textbf{0.921}\\ \hline 
\end{tabular}
%\caption{Comparison of different weighting approaches on the WV3 reduced-resolution dataset.}
\label{table:diff_weight}
\vspace{-0.5cm}
\end{table}

\begin{figure}[t]
   \centering
   \includegraphics[width=1.0\linewidth]{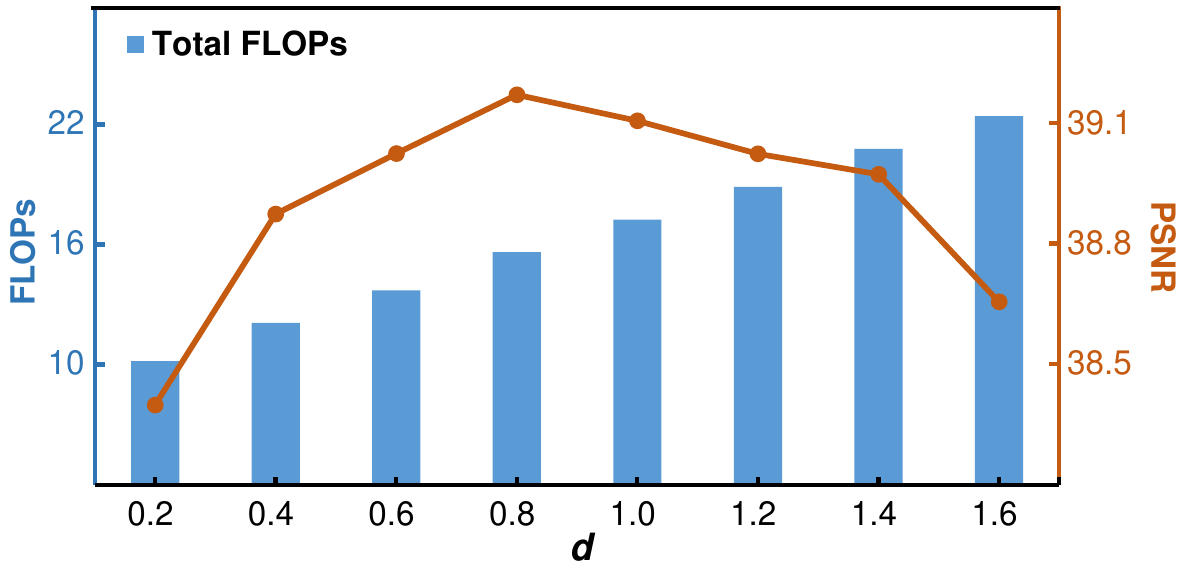}
   \vspace{-0.5cm}
   \caption{FLOPs (K) and PSNR (dB) with \( d \) represented as a fraction of \( n \), both are variables in \cref{fig:weighting}.}
   \label{fig:H}
   \vspace{-0.4cm}
\end{figure}

\noindent\textbf{Impact of Intermediate Layer Size:} 
This section explores the impact of the intermediate layer size \( d \), a key variable in our method. Experiments were conducted using LAGNet~\cite{LAGConv} as the backbone on the WV3 reduced-resolution dataset. 
In this analysis, \( d \) is varied in IFW while keeping CFW fixed, with its FLOPs constant at 90, contributing only a small portion to the total computation.
As shown in Fig.~\ref{fig:H}, the overall FLOPs increase as \( d \) becomes larger. In terms of performance, when \( d \) is too small, the network struggles to capture complex feature correlations, resulting in lower PSNR values. As \( d \) increases, PSNR improves, indicating enhanced image quality. However, when \( d \) exceeds 0.8\( n \), the network becomes overly complex, leading to overfitting and a subsequent drop in PSNR. 

\begin{table}[t]
\centering
\caption{Ablation experiment on WV3 reduced-resolution dataset.}
\vspace{-0.1cm}
\setlength{\tabcolsep}{4pt} % 默认值为6pt，调整为4pt以缩短列间距
\centering
\begin{tabular}{cccccc}
\hline
\vspace{-3pt}
\multirow{2}{*}{\textbf{IFW}} & \multirow{2}{*}{\textbf{CFW}} & \multicolumn{4}{c}{\textbf{WV3}}                                    \\ 
\cmidrule(lr){3-6}
                                       &                                      & \textbf{PSNR↑} & \textbf{SAM↓}  & \textbf{ERGAS↓} & \textbf{Q8↑}   \\ \hline
\ding{55}                          & \ding{55}                                      & 38.592       & 3.103      & 2.291       & 0.910      \\
\ding{51}                             & \ding{55}                                      & 39.028          & 2.972          & 2.183           & 0.917          \\
\ding{55}                            & \ding{51}                                    & 38.923 & 2.990 & 2.210 & 0.918 \\
\ding{51}                             & \ding{51}                                    & \textbf{39.170} & \textbf{2.913} & \textbf{2.145} & \textbf{0.921} \\ \hline
\end{tabular}
\label{table:ablation}
\vspace{-0.6cm}
\end{table}

\vspace{3pt}
\noindent\textbf{Complexity Analysis}:
The additional computational complexity of ADWM, compared to the original network, mainly stems from generating the covariance matrix.
IFW has a theoretical complexity of \(O(C^2 \cdot (H \cdot W))\), where \(H\) and \(W\) are the spatial dimensions, and \(C\) is the number of channels.
CFW has a theoretical complexity of \(O(N^2 \cdot C)\), where \(N\) is the number of feature maps being processed.
In total, the complexity of our module is \( O(N \cdot (H \cdot W) \cdot C^2 + N^2 \cdot C) \), indicating that the main computational cost is driven by the number of channels \(C\) and the spatial dimensions \(H\) and \(W\).

\vspace{-2pt}
\subsection{Ablation Study}
\vspace{-1pt}
We conducted ablation studies with LAGNet~\cite{LAGConv} on the WV3 dataset. In the second row of \cref{table:ablation}, replacing CFW's dynamic weighting with equal weights causes a performance drop, highlighting the importance of dynamic adjustment. In the third row, omitting IFW caused a significant decline, underscoring its role in handling feature heterogeneity and redundancy. However, performance still exceeded the first row, demonstrating CFW's effectiveness in integrating shallow and deep features.
\vspace{-5pt}
\section{Conclusion}
\vspace{-5pt}
\label{sec:conclusion}
In this paper, we use the covariance matrix to model feature heterogeneity and redundancy, introducing CACW to capture these correlations and generate weights for effective adjustment. Building on CACW, we propose the adaptive dual-level weighting mechanism (ADWM), which includes Intra-Feature Weighting (IFW) and Cross-Feature Weighting (CFW). ADWM significantly enhances a wide range of existing deep learning methods, and its effectiveness is thoroughly demonstrated through extensive experiments.

% In this paper, we use the covariance matrix to model feature heterogeneity and redundancy, and propose CACW, which leverages the covariance matrix to capture these correlations and generate weights for adjustment. Building upon CACW, we propose a general adaptive dual-level weighting mechanism (ADWM) including Intra-Feature Weighting (IFW) and Cross-Feature Weighting (CFW). ADWM enhances a wide range of existing deep learning methods, and its effectiveness is thoroughly demonstrated through extensive experiments.

\noindent\textbf{Acknowledgement:}
This work was supported in part by the NSFC (12271083, 62273077), Sichuan Province's Science and Technology Empowerment for Disaster Prevention, Mitigation, and Relief Project (2025YFNH0001), and the Natural Science Foundation of Sichuan Province (2024NSFJQ0013).

{\small
\bibliographystyle{ieeenat_fullname}
\bibliography{11_references}
}

\ifarxiv \clearpage \appendix \begin{abstract}
The supplementary materials provide additional insights into the ADWM proposed in our paper. We present an in-depth exploration of how our method leverages covariance matrices and draws connections to PCA, highlighting its distinctions from prior approaches. Key variables influencing performance are thoroughly examined. The implementation approach, datasets, and training details are also provided. Finally, we present additional experimental results on visual analysis, comparison with SOTA methods, and the generality experiment. Code will be provided after acceptance.
\end{abstract}

\section{Analysis on Covariance Matrix and PCA}
Many prior works~\cite{wang2020eca,ling2019Sra,Hu2013senet} have explored adaptive weighting to enhance feature representation. Our method introduces two key innovations: 1) leveraging covariance matrix correlations to explicitly capture feature heterogeneity and redundancy; and 2) both IFW and CFW have conceptual links to PCA, providing stronger theoretical support.

\noindent{\bf Covariance matrix}. 
Covariance matrices are symmetric, with off-diagonal elements \( C_{ij} \) denoting the covariance between features \( i \) and \( j \). High absolute covariance values indicate strong linear correlations, while low values suggest weaker dependencies. 
In our method, covariance captures both feature redundancy and heterogeneity. High covariance often implies redundancy, as features convey overlapping information. For example, two features strongly correlated with shared attributes (e.g., brightness or texture) result in high covariance. In contrast, low covariance reflects heterogeneity, where features provide distinct information, such as spectral properties versus spatial textures. 
This approach offers a significant advantage over traditional weighting mechanisms, which rely on simple global statistics or local operations and fail to model complex global relationships among features.
By explicitly modeling these relationships using the covariance matrix, \textit{our method inherently incorporates non-local properties, quantifying the dependencies between features. As a result, weights are adaptively adjusted to enhance features with high heterogeneity, while suppressing those with high redundancy. This not only reduces redundant information but also significantly improves the overall informativeness of the feature representations.}

\noindent{\bf Link Between CFW and PCA}. 
In PCA, the eigenvectors \( \mathbf{v}_i \) represent the principal directions of variation in the data. Similarly, our CACW generates weights \( \beta \) that form a basis to highlight important features and suppress redundancy, providing a clear and interpretable theoretical foundation. The relationship between CFW and PCA is particularly notable, as it parallels PCA's projection operation, where the original data matrix \( X \) is transformed using the eigenvector matrix \( P \), formulated as:
\begin{equation}
\label{equ:transform}
Y = P^T X,
\end{equation}
where \( Y \) represents the reduced-dimensional data. The pointwise weighting and summation process in CFW can also be rewritten in a matrix multiplication form, which corresponds to \cref{equ:transform} in structure. 
The weighting process in our method can be formulated as:
\begin{equation}
\hat{F} = (\textrm{softmax}(\beta))^T \tilde{F},
\end{equation}
where \( \hat{F} \) represents the aggregated critical information from all intermediate results, effectively reducing redundancy and capturing the essence of the feature space.
However, unlike PCA's global dimensionality reduction using a fixed orthogonal basis, CFW dynamically learns and applies task-specific weights \( \beta \). This adaptive aggregation tailors feature representation to the task, enhancing relevant features and overall representation.

\noindent{\bf Link Between IFW and PCA}. 
The IFW also demonstrates a conceptual link to PCA, though with a key distinction. Instead of performing a matrix multiplication for dimensionality reduction, IFW employs a channel-wise pointwise multiplication between the generated weights and the original features. The weighting process is formulated as follows:
\begin{equation}
\tilde{F_i} = F_i \odot \alpha_i,
\vspace{-3pt}
\end{equation}
where \( \odot \) denotes element-wise multiplication.
Unlike the global orthogonal projection in PCA represented by \cref{equ:transform}, IFW independently scales each feature dimension through the generated weights. This can be seen as a simplified form of projection that preserves the original feature basis while optimizing the distribution of information. By selectively amplifying important feature channels and suppressing redundant ones, IFW refines the feature representation without requiring a global basis transformation.

\begin{figure*}[t]
  \centering
  \includegraphics[width=1.0\linewidth]{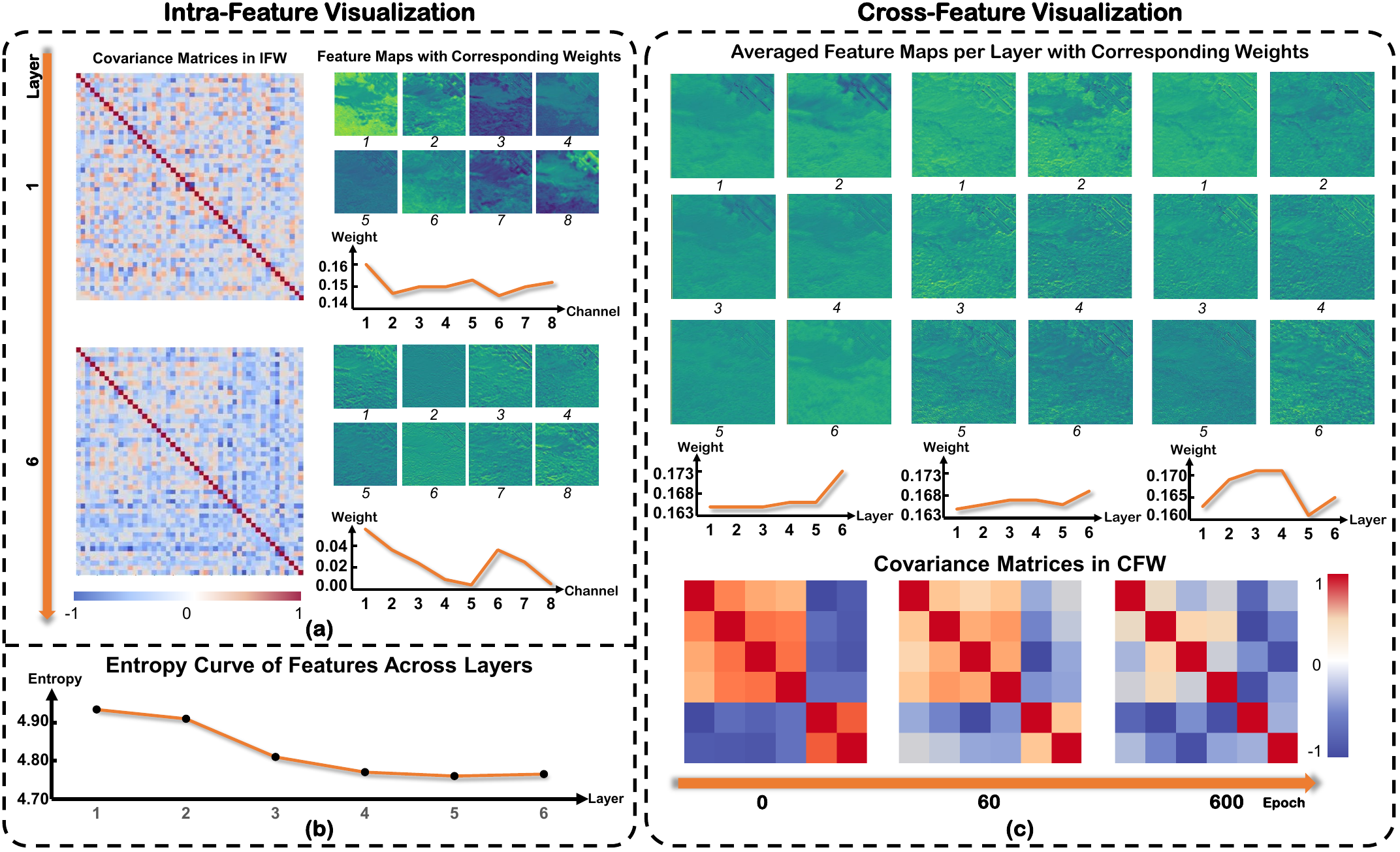}
  \caption{
Visualization of covariance matrices, weights, and feature representations in IFW and CFW. In (a), intra-feature covariance matrices, channel weights, and corresponding feature maps across various layers are shown, with channels that are multiples of six selected for clarity. In (b), entropy variations of features across layers are displayed. In (c), cross-feature covariance matrices, layer weights, and average feature maps across different training epochs are illustrated.
}
  \label{fig:suppl_visual}
  \vspace{-0.5cm}
\end{figure*}

\begin{figure}[t]
   \centering
   \includegraphics[width=1.0\linewidth]{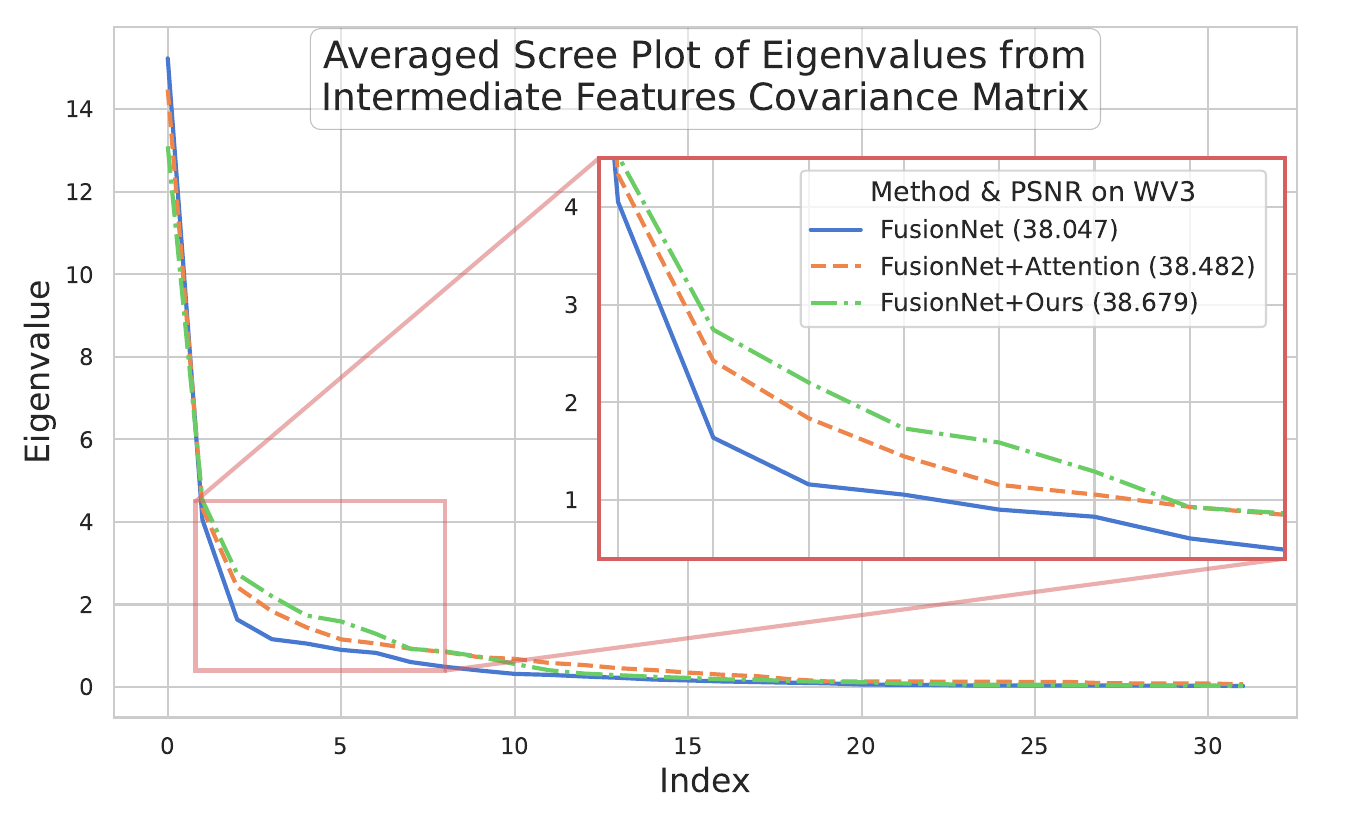}
   \caption{
   Scree Plot to illustrate the differences in redundancy. The smoother the curve, the lower the redundancy.}
   \label{fig:suppl_pnp}
   \vspace{-0.5cm}
\end{figure}

\subsection{Visual Analysis}
\subsubsection{Intra-Feature Visualization}
\vspace{-4pt}
\noindent\textbf{Change of Covariance Matrix}:
As shown in \cref{fig:suppl_visual} (a), 
In shallower layers, the overall color of the matrix is lighter, indicating more diverse information across channels, but it becomes darker with increasing depth, reflecting high redundancy.
Additionally, as shown in \cref{fig:suppl_visual} (b), the entropy curve illustrates the level of information diversity across layers, with a decreasing trend indicating increased redundancy as the depth increases.

\vspace{2pt}
\noindent\textbf{Changes of Channel Weight}:
The line charts within the feature maps illustrate the weights assigned to each channel. In the shallow layers, the weights are relatively uniform (0.141 to 0.155), indicating minimal differentiation among low-level features that do not require emphasis. In deeper layers, however, the weights show greater variance (0.004 to 0.055), with some channels emphasized and others minimized. The smaller weight values in deeper layers reflect the larger magnitude of deep-layer features. This distribution enables the model to capture foundational information in the shallow layers and selectively focus on important structural details in deeper layers.

\vspace{2pt}
\noindent\textbf{Redundancy Visualization}:
To further demonstrate that our method improves by reducing feature redundancy and enhancing heterogeneity, we additionally visualized the results using FusionNet~\cite{FusionNet} on the WV3 reduced-resolution dataset. The analysis is the same as in Sec. 3.4 of the main text.

\vspace{-10pt}
\subsubsection{Cross-Feature Visualization}
\vspace{-5pt}
\noindent\textbf{Change of Covariance Matrix}:
As illustrated in \cref{fig:suppl_visual} (c), The covariance matrix generated by CFW shows significant evolution over the course of training. 
In the early epochs, the blue areas indicate negative correlations between shallow and deep features, while the red areas reveal positive correlations within the shallow and deep features themselves. Both types of correlations reflect redundancy in the feature representation.
As training progresses, the overall color of the covariance matrix lightens, reflecting a gradual reduction in inter-feature correlation. This trend indicates increased feature diversity and a decline in redundant information as training advances.

\vspace{2pt}
\noindent\textbf{Changes of Layer Weight}:
As training progresses, layer weights adjust gradually, indicating the model’s adaptive tuning of each layer's impact. These shifts reflect the model’s refinement to align layer contributions with task demands, enhancing its ability to leverage diverse features and optimize performance throughout learning.

\section{Implementation Details of Plug-and-Play}  

This section demonstrates how our method can be seamlessly integrated into various existing approaches. 
Our ADWM serves as a plug-and-play mechanism.
When comparing with SOTA methods, we incorporated the ADWM module into the classic LAGNet~\cite{LAGConv} as our proposed method. To further validate the generality of our approach, we integrated three baseline methods into our ADWM framework: FusionNet~\cite{FusionNet} and LGPNet~\cite{zhao2023lgpconv}. 
The U2Net~\cite{U2Net} has already been presented in the main text, Fig. 6.
As shown in Fig.~\ref{fig:pnp},
all methods retain their original frameworks, with our dual-level weighting module integrated only into the intermediate continuous feature extraction blocks.

\section{Datasets}  
\begin{figure}[t]
   \centering
   \includegraphics[width=1.0\linewidth]{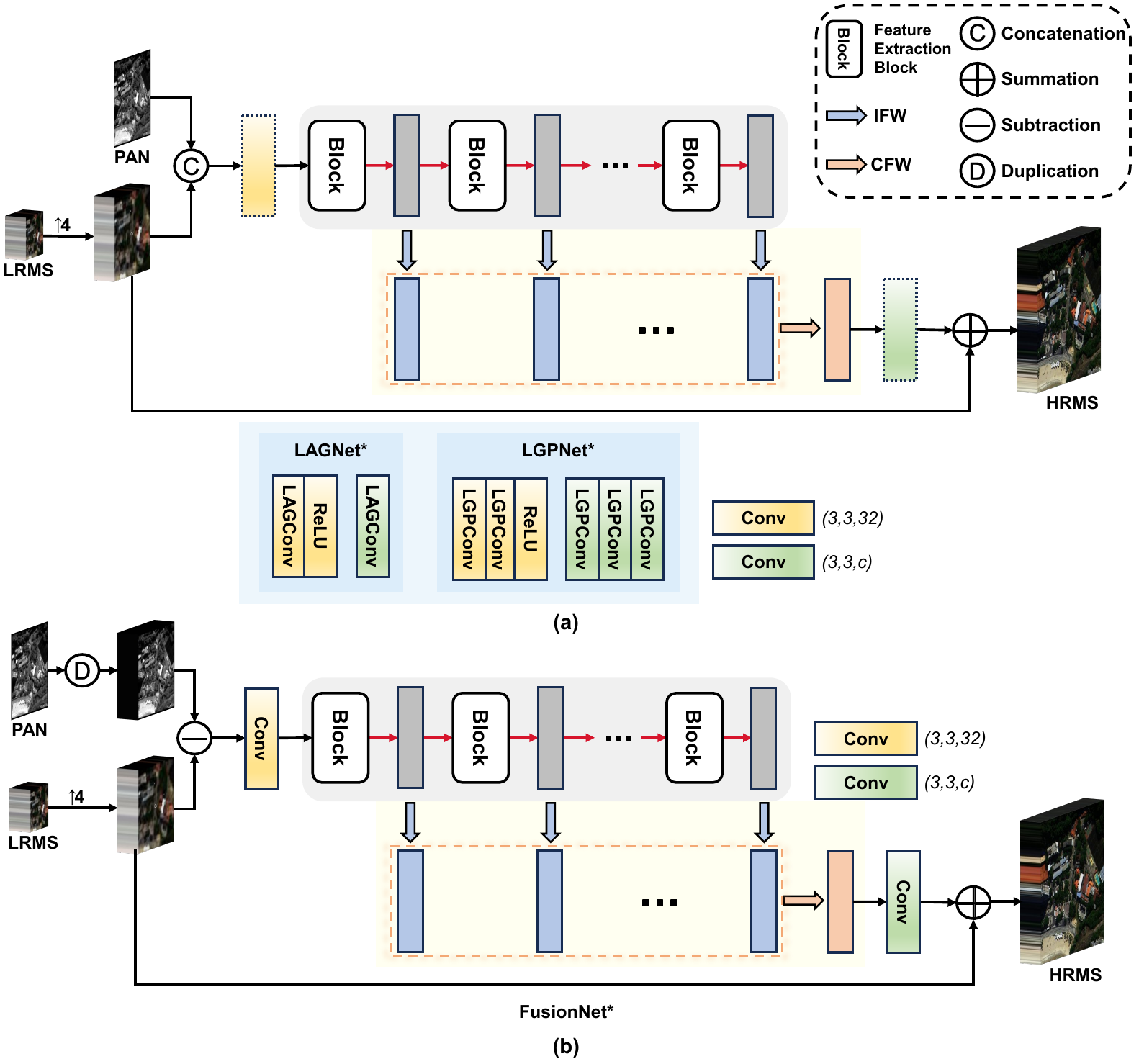}
   \caption{
   Illustrations of how our proposed module is integrated into various methods in a plug-and-play manner. (a) Integration into LAGNet~\cite{LAGConv} and LGPNet~\cite{zhao2023lgpconv}, which share the same framework but differ in their feature extraction blocks. The unique designs of the input and output stages are highlighted at the bottom. (b) Integration into FusionNet~\cite{FusionNet}.
   }
   \label{fig:pnp}
   \vspace{-0.5cm}
\end{figure}

We utilized datasets derived from the WorldView-3 (WV3), QuickBird (QB), and GaoFen-2 (GF2) satellites for our experiments. These datasets consist of image patches extracted from remote sensing imagery, which are separated into training and testing subsets. The training data includes image triplets of PAN, LRMS, and GT obtained through downsampling-based simulation, with respective dimensions of $64 \times 64$, $16 \times 16 \times C$, and $64 \times 64 \times C$. For the WV3 dataset, the training set contains approximately 10,000 samples with eight channels ($C=8$). Similarly, the QB training set consists of about 17,000 samples with four channels ($C=4$), while the GF2 dataset includes 20,000 samples with four channels ($C=4$). The reduced-resolution test set for each satellite is composed of 20 PAN/LRMS/GT image triplets with a variety of representative land cover types. These images, simulated via downsampling, have dimensions of $256 \times 256$, $64 \times 64 \times C$, and $256 \times 256 \times C$, respectively. For the full-resolution testing phase, the dataset comprises 20 PAN/LRMS image pairs with sizes of $512 \times 512$ and $128 \times 128$. The datasets and processing procedures were obtained from the PanCollection repository~\cite{dengMachineLearningPansharpening2022}.

\begin{table*}[t]
\centering
\caption{Comparisons on WV3, QB, and GF2 full-resolution datasets, each with 20 samples. Best: \textbf{bold}, and second-best: \underline{underline}.}
\vspace{-0.1cm}
\begin{tabular}{cc@{\hskip 3pt}c@{\hskip 3pt}cc@{\hskip 3pt}c@{\hskip 3pt}cc@{\hskip 3pt}c@{\hskip 3pt}c}
\hline
\vspace{-3pt}
\multirow{2}{*}{\textbf{Methods}} & \multicolumn{3}{c}{\textbf{ WV3 }} & \multicolumn{3}{c}{\textbf{QB}} & \multicolumn{3}{c}{\textbf{GF2}}\\ 
\cmidrule(lr){2-4}  \cmidrule(lr){5-7}  \cmidrule(lr){8-10}  
 & \textbf{D$_\lambda\downarrow$}& \textbf{D$_s\downarrow$}& \textbf{HQNR$\uparrow$}& \textbf{D$_\lambda\downarrow$}& \textbf{D$_s\downarrow$}& \textbf{HQNR$\uparrow$}& \textbf{D$_\lambda\downarrow$}& \textbf{D$_s\downarrow$}& \textbf{HQNR$\uparrow$}\\ \hline
MTF-GLP-FS~\cite{MTF-GLP-FS} & 0.020& 0.063& 0.919& 0.047& 0.150& 0.810& 0.035& 0.143& 0.828\\
BDSD-PC~\cite{BDSD-PC} & 0.063& 0.073& 0.870& 0.198& 0.164& 0.672& 0.076& 0.155& 0.781\\
TV~\cite{TV} & 0.023& 0.039& 0.938& 0.055& 0.100& 0.850& 0.055& 0.112& 0.839\\
\hline
PNN~\cite{PNN} & 0.021& 0.043& 0.937& 0.058& 0.062& 0.884& 0.032& 0.094& 0.877\\
PanNet~\cite{Yang2017PanNet} & \textbf{0.017}& 0.047& 0.937& \textbf{0.043}& 0.114& 0.849& \textbf{0.018}& 0.080& 0.904\\
DiCNN~\cite{hePansharpeningDetailInjection2019} & 0.036& 0.046& 0.920& 0.095& 0.107& 0.809& 0.037& 0.099& 0.868\\
FusionNet~\cite{FusionNet} & 0.024& 0.037& 0.940& 0.057& 0.052& 0.894& 0.035& 0.101& 0.867\\
LAGNet~\cite{LAGConv} & 0.037& 0.042& 0.923& 0.086& 0.068& 0.852& 0.028& 0.079& 0.895\\
LGPNet~\cite{zhao2023lgpconv} & 0.022& 0.039& 0.940& 0.074& 0.061& 0.870& 0.030& 0.080& 0.892\\
PanMamba~\cite{He2024PanMambaEP} & \underline{0.018}& 0.053& 0.930& \underline{0.049}& \underline{0.044}& \underline{0.910}&  0.023& 0.057& 0.921\\
% ZS-Pan~\cite{caoif2024} & 0.021& \underline{0.032}& \underline{0.947}& 0.058& 0.046& 0.898& 0.023& \underline{0.055}& \underline{0.923}\\
\textbf{Proposed} & 0.024& \textbf{0.029}& \textbf{0.948}& 0.064& \textbf{0.024}& \textbf{0.914}& \underline{0.022}& \textbf{0.052}& \textbf{0.928}\\
\hline
\end{tabular}
\label{tab:all_full}
%\vspace{-0.3cm}
\end{table*}

\begin{table*}[t]
\centering
\caption{Comparisons on WV3, QB, and GF2 datasets with 20 reduce-resolution samples, respectively. Methods marked with * represent the corresponding method enhanced with our ADWM module without any further changes. The best results in each column are \textbf{bolded}.}
\vspace{-0.1cm}
	\begin{tabular}{c@{\hskip 5pt}c@{\hskip 7pt}c@{\hskip 4pt}c@{\hskip 5pt}c@{\hskip 7pt}c@{\hskip 7pt}c@{\hskip 4pt}c@{\hskip 5pt}c@{\hskip 7pt}c@{\hskip 7pt}c@{\hskip 4pt}c@{\hskip 5pt}c}
		\hline
        \vspace{-3pt}
		\multirow{2}{*}{\textbf{Method}}  & \multicolumn{4}{c}{\textbf{WV3}} & \multicolumn{4}{c}{\textbf{QB}} & \multicolumn{4}{c}{\textbf{GF2}}                                 \\ 
        \cmidrule(lr){2-5}  \cmidrule(lr){6-9}  \cmidrule(lr){10-13}  
		              & \textbf{PSNR$\uparrow$} & \textbf{SAM$\downarrow$} & \textbf{ERGAS$\downarrow$} & \textbf{Q8$\uparrow$} 
        & \textbf{PSNR$\uparrow$} & \textbf{SAM$\downarrow$} & \textbf{ERGAS$\downarrow$} & \textbf{Q4$\uparrow$} 
        & \textbf{PSNR$\uparrow$} & \textbf{SAM$\downarrow$} & \textbf{ERGAS$\downarrow$} & \textbf{Q4$\uparrow$} \\ \hline
		\textbf{FusionNet}~\cite{FusionNet}               &                                38.047 & 3.324 & 2.465 & 0.904 & 37.540 & 4.904 & 4.156 & 0.925 & 39.639 & 0.974 & 0.988 & 0.964 
\\
		\textbf{FusionNet*}              &                                \textbf{38.679}& \textbf{3.097}& \textbf{2.249}& \textbf{0.909}& \textbf{38.271}& \textbf{4.654}& \textbf{3.795}& \textbf{0.933}& \textbf{41.649}& \textbf{0.832}& \textbf{0.769}& \textbf{0.975}\\ \hline
        \textbf{LGPNet}~\cite{zhao2023lgpconv}                  &                                38.147& 3.270& 2.422& \textbf{0.902}& 36.443& 4.954& 4.777& 0.915& 41.843& 0.845& 0.765& 0.976
\\
		\textbf{LGPNet*} &                                  \textbf{38.400}& \textbf{3.241}& \textbf{2.330}& \textbf{0.902}& \textbf{38.309}& \textbf{4.666}& \textbf{3.773}& \textbf{0.932}& \textbf{42.230}& \textbf{0.828}& \textbf{0.714}& \textbf{0.978}\\
	\hline % 添加此行
    \textbf{U2Net}~\cite{U2Net}                    & 39.117 & 2.888 & 2.149 & 0.920 & 38.065 & 4.642 & 3.987 & 0.931 & 43.379 & 0.714 & 0.632 & 0.981 \\ 
		\textbf{U2Net*}                  &                               \textbf{39.234}& \textbf{2.864}& \textbf{2.143}& \textbf{0.922}& \textbf{38.762}& \textbf{4.432}& \textbf{3.691}& \textbf{0.939}& \textbf{43.901}& \textbf{0.670}& \textbf{0.592}& \textbf{0.986}\\ \hline
    \end{tabular}

\label{table:general_reduce}
\vspace{-0.3cm}
\end{table*}

\section{Training Details}
When comparing with SOTA methods, the training of ADWM on the WV3 dataset was conducted using the $\ell_1$ loss function and the Adam optimizer. The batch size was set to 64, with an initial learning rate of $2 \times 10^{-3}$, decaying to half its value every 150 steps. The training process spanned 500 epochs. The network architecture used 48 channels in the hidden layers, and the intermediate layer size \( d \) in CACW was set to 0.8\( n \).
For the QB dataset, the training process lasted 200 epochs, with the intermediate layer size \( d \) in CACW set to 0.6\( n \), while all other settings remained consistent with those used for the WV3 dataset. For the GF2 dataset, all settings were identical to the WV3 configuration.
In general experiments, all other training settings followed those specified in the original papers.

\section{Additional Results}
\noindent{\bf Visual Analysis}: 
To provide a more detailed illustration of the channel weights generated in IFW, we present the complete results of the first and sixth layers on one picture of GF2 reduced-resolution datasets, including all feature maps and their corresponding weights for each channel, as shown in Fig.~\ref{fig:sup_visual}.

\noindent{\bf Comparison with SOTA methods}: 
\cref{tab:all_full} showcases a comprehensive comparison of our method with state-of-the-art approaches across three datasets on full-resolution images. The HQNR metric~\cite{hqnr}, which improves upon the QNR metric, evaluates both spatial and spectral consistency, offering a comprehensive reflection of the image-fusion effectiveness of different methods. It is widely regarded as one of the most important metrics for full-resolution datasets.  
Our method achieves the highest HQNR on all three datasets.
Additionally, in \cref{fig:sup_eps_qb_sota,fig:sup_eps_wv3_sota,fig:sup_hot_gf2_sota,fig:sup_hot_qb_sota,fig:sup_hot_wv3_sota}, we provide visual comparisons of the outputs generated by various methods on sample images from the WV3, QB, and GF2 datasets. 

\noindent{\bf Generality Experiment}:  
\cref{table:general_reduce} presents the results of the generality experiment conducted across three datasets on reduced-resolution images. In \cref{fig:sup_eps_gf2_gen,fig:sup_eps_qb_gen,fig:sup_eps_wv3_gen,fig:sup_hot_gf2_gen,fig:sup_hot_qb_gen}, we showcase visual output comparisons for some methods before and after incorporating our ADWM module, using sample images from the WV3, QB, and GF2 datasets.

\begin{figure*}[ht]
   \centering
   \includegraphics[width=1\linewidth]{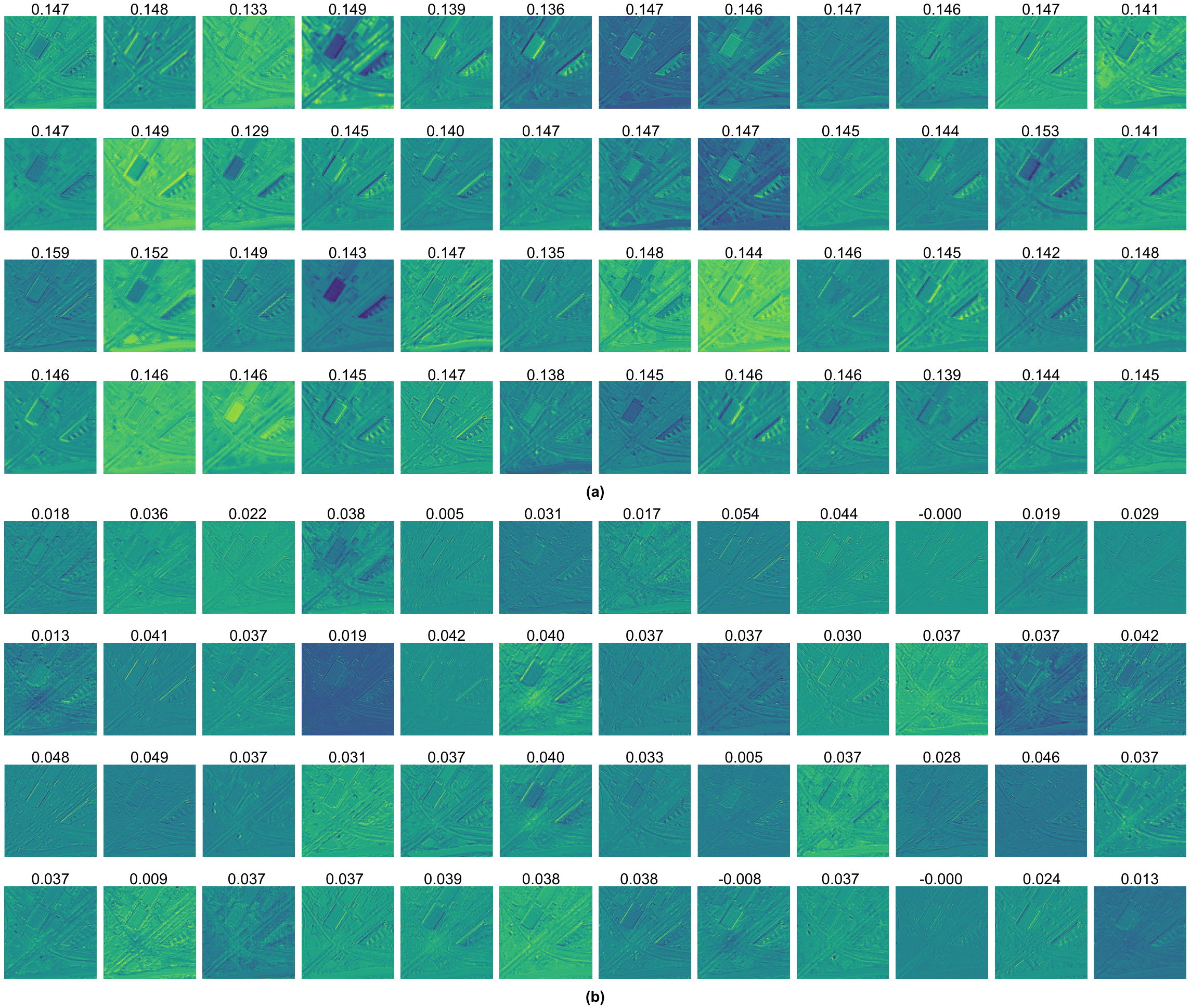}

   \caption{Channel weights and corresponding feature maps in IFW: (a) Results from the first layer, where each feature map is annotated with its corresponding weight. (b) Results from the sixth layer.}
   \label{fig:sup_visual}
   \vspace{-0.5cm}
\end{figure*}

\begin{figure*}[ht]
   \centering
   \includegraphics[width=1\linewidth]{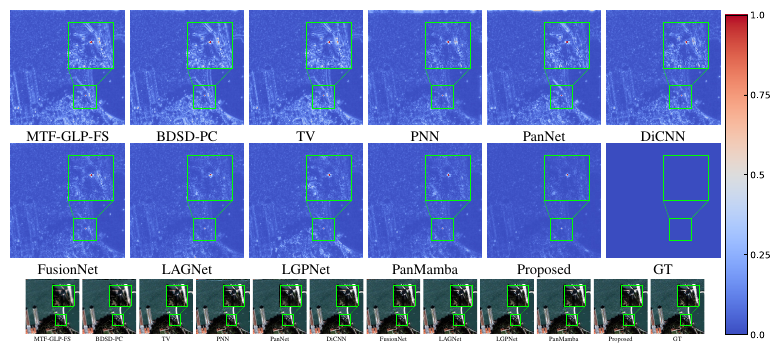}

   \caption{The residuals (Top) and visual results (bottom) of all compared approaches on the QB reduced-resolution dataset.}
   \label{fig:sup_eps_qb_sota}
   %\vspace{-0.5cm}
\end{figure*}

\begin{figure*}[ht]
   \centering
   \includegraphics[width=1\linewidth]{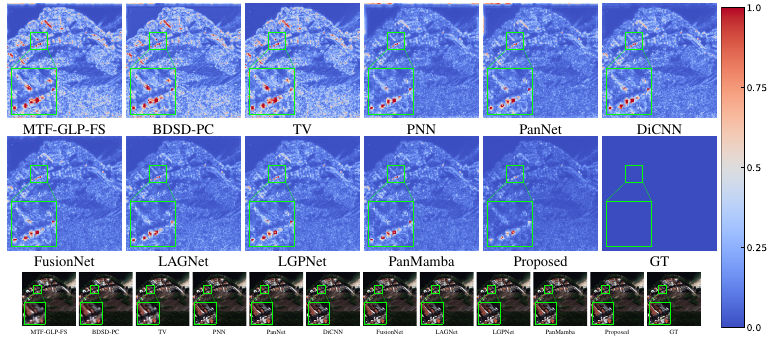}

   \caption{The residuals (Top) and visual results (bottom) of all compared approaches on the WV3 reduced-resolution dataset.}
   \label{fig:sup_eps_wv3_sota}
   \vspace{-0.5cm}
\end{figure*}

\begin{figure*}[ht]
   \centering
   \includegraphics[width=1\linewidth]{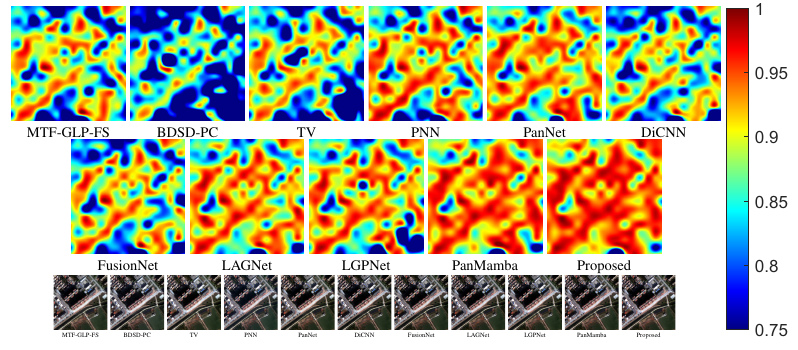}
   \caption{The HQNR maps (Top) and visual results (bottom) of all compared approaches on the GF2 full-resolution dataset.}
   \label{fig:sup_hot_gf2_sota}
   \vspace{-0.5cm}
\end{figure*}

\begin{figure*}[ht]
   \centering
   \includegraphics[width=1\linewidth]{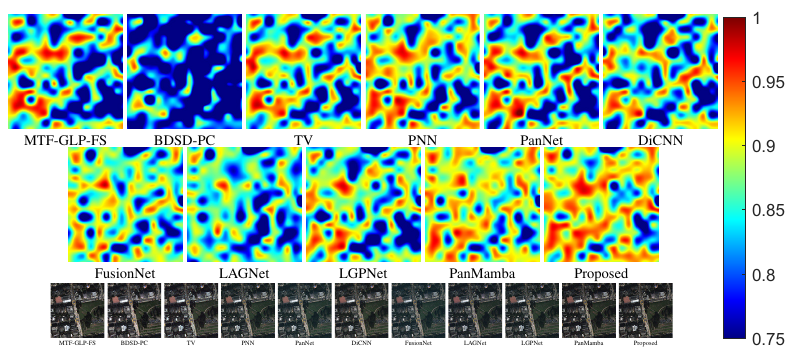}
   \caption{The HQNR maps (Top) and visual results (bottom) of all compared approaches on the QB full-resolution dataset.}
   \label{fig:sup_hot_qb_sota}
   \vspace{-0.5cm}
\end{figure*}

\begin{figure*}[ht]
   \centering
   \includegraphics[width=1.0\linewidth]{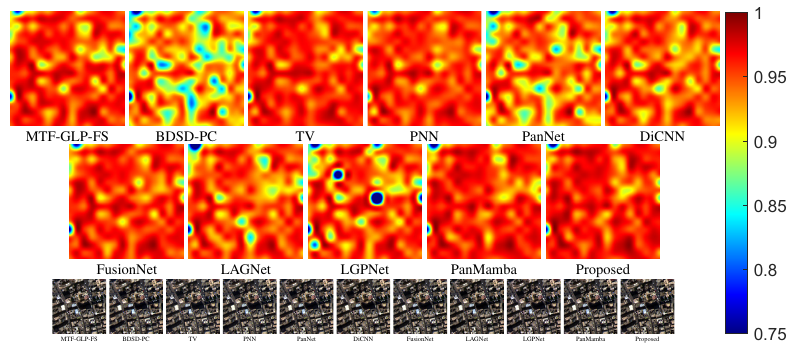}
%\vspace{-0.2cm}
   \caption{The HQNR maps (Top) and visual results (bottom) of all compared approaches on the WV3 full-resolution dataset.}
   \label{fig:sup_hot_wv3_sota}
   %\vspace{-0.5cm}
\end{figure*}

\begin{figure*}[ht]
   \centering
   \includegraphics[width=1.0\linewidth]{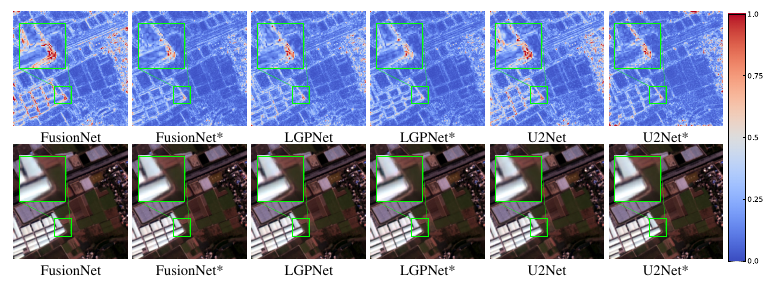}

   \caption{The residuals (Top) and visual results (Bottom) of all evaluated general methods on the GF2 reduced-resolution dataset.}
   \label{fig:sup_eps_gf2_gen}
   %\vspace{-0.5cm}
\end{figure*}

\begin{figure*}[ht]
   \centering
   \includegraphics[width=1.0\linewidth]{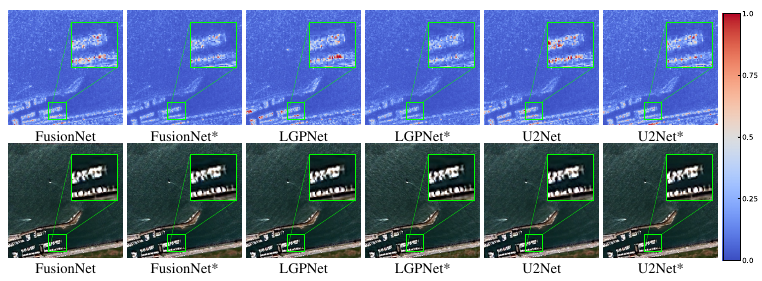}

   \caption{The residuals (Top) and visual results (Bottom) of all evaluated general methods on the QB reduced-resolution dataset.}
   \label{fig:sup_eps_qb_gen}
   \vspace{-0.5cm}
\end{figure*}

\begin{figure*}[ht]
   \centering
   \includegraphics[width=1.0\linewidth]{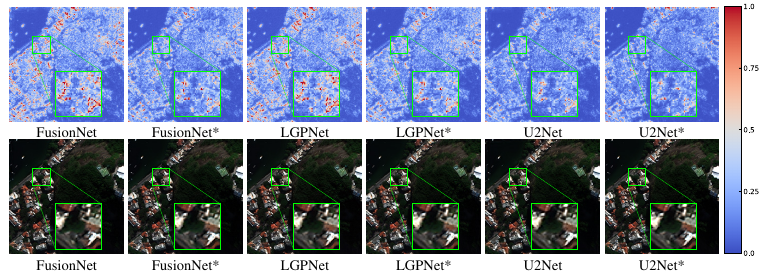}

   \caption{The residuals (Top) and visual results (Bottom) of all evaluated general methods on the WV3 reduced-resolution dataset.}
   \label{fig:sup_eps_wv3_gen}
   \vspace{-0.5cm}
\end{figure*}

\begin{figure*}[ht]
   \centering
   \includegraphics[width=1.0\linewidth]{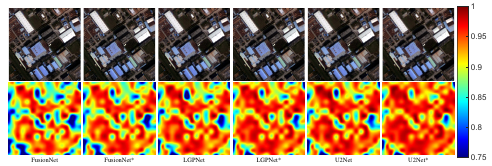}

   \caption{The visual results (Top) and HQNR maps (Bottom) of all evaluated general methods on the GF2 full-resolution dataset.}
   \label{fig:sup_hot_gf2_gen}
   \vspace{-0.5cm}
\end{figure*}

\begin{figure*}[ht]
   \centering
   \includegraphics[width=1.0\linewidth]{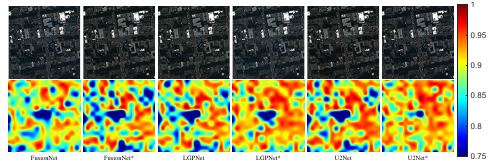}

   \caption{The visual results (Top) and HQNR maps (Bottom) of all evaluated general methods on the QB full-resolution dataset.}
   \label{fig:sup_hot_qb_gen}
   \vspace{-0.5cm}
\end{figure*}
 \fi

\end{document}